\documentclass[]{fairmeta}

\title{\mvdustthreerp: Single-Stage Scene Reconstruction from Sparse Views In 2 Seconds}

\author[1,2]{Zhenggang Tang}
\author[1]{Yuchen Fan}
\author[1]{Dilin Wang}
\author[1]{Hongyu Xu}
\author[1]{Rakesh Ranjan}
\author[2]{Alexander Schwing}
\author[1,\dagger]{Zhicheng Yan}

\affiliation[1]{Meta Reality Labs}
\affiliation[2]{University of Illinois Urbana-Champaign}
\contribution[\dagger]{project lead}

\abstract{
Recent sparse multi-view scene reconstruction advances like \dst\ and \mastthreer no longer require camera calibration and camera pose estimation. However, they only process a pair of views at a time to infer pixel-aligned pointmaps. When dealing with more than two views, a combinatorial number of error prone pairwise reconstructions are usually followed by an expensive global optimization, which often fails to rectify the 
pairwise reconstruction errors.  To handle more views, reduce errors, and improve inference time, we propose the fast single-stage feed-forward network \mvdustthreer. At its core are multi-view decoder blocks which exchange information across any number of views while considering one reference view. To make our method  robust to reference view selection, we further propose \mvdustthreerp, which employs cross-reference-view blocks to fuse information across different reference view choices. To further enable novel view synthesis, we extend both by adding and jointly training Gaussian splatting heads. Experiments on multi-view stereo reconstruction, multi-view pose estimation, and novel view synthesis confirm that our methods improve significantly upon prior art. Code will be released.
}

\correspondence{\email{zyan3@meta.com}}
\metadata[Website]{\url{https://mv-dust3rp.github.io/}}
\usepackage{hyperref}
\usepackage{url}
\usepackage[T1]{fontenc}    %
\usepackage{url}            %
\usepackage{booktabs}       %
\usepackage{amsfonts}       %
\usepackage{nicefrac}       %
\usepackage{microtype}      %
\usepackage{leftindex}
\usepackage{comment}
\usepackage[inkscapelatex=false]{svg}

\usepackage{epsfig, xspace, enumitem}
\usepackage{cases}
\usepackage{graphicx, amsmath, amssymb, caption, subcaption, multirow, overpic, textpos, pifont, adjustbox}

\usepackage[utf8x]{inputenc}
\makeatletter
\@namedef{ver@everyshi.sty}{}
\makeatother
\usepackage{pgf, tikz, pgfplots}
\usepackage{makecell}
\usepackage{pgf-pie}
\usepackage{float}
\usepackage{wrapfig}
\usepackage{colortbl}
\usepackage[title]{appendix}

\definecolor{citecolor}{HTML}{0071BC}
\definecolor{linkcolor}{HTML}{ED1C24}
\definecolor{acceptcolor}{HTML}{74C219}
\definecolor{rejectcolor}{HTML}{DE1616}
\definecolor{qcolor}{HTML}{536872}
\definecolor{demphcolor}{RGB}{100,100,100}
\definecolor{brightlavender}{rgb}{0.75, 0.58, 0.89}
\definecolor{palered}{rgb}{1.00, 0.70, 0.70}
\definecolor{palegreen}{rgb}{0.73, 0.96, 0.67}
\definecolor{paleblue}{rgb}{0.69, 0.84, 1.00}
\definecolor{paleorange}{rgb}{1.00, 0.86, 0.73}
\definecolor{palepurple}{rgb}{0.92, 0.85, 1.00}
\definecolor{paleyellow}{rgb}{1.00, 1.00, 0.50}

\def\dst{DUSt3R}

\makeatletter
\DeclareRobustCommand\onedot{\futurelet\@let@token\@onedot}
\def\@onedot{\ifx\@let@token.\else.\null\fi\xspace}

\def\eg{\emph{e.g}\onedot} 
\def\ie{\emph{i.e}\onedot} 
 
\def\etc{\emph{etc}\onedot}

\makeatother

\newcommand{\lgr}[1]{\textcolor{lightgray}{#1}}
\newcommand{\dustthreer}[0]{DUSt3R\xspace}
\newcommand{\mastthreer}[0]{MASt3R\xspace}
\newcommand{\mvdustthreer}[0]{MV-DUSt3R\xspace}
\newcommand{\mvdustthreerp}[0]{MV-DUSt3R+\xspace}

\newcommand{\crossrefview}[0]{Cross-Reference-View\xspace}
\newcommand{\scannet}[0]{ScanNet}
\newcommand{\scannetspace}[0]{ScanNet }

\newcommand{\lconf}{\mathcal{L}_{\text{conf}}} %
\newcommand{\lrender}{\mathcal{L}_{\text{render}}} %
\newcommand{\lall}{\mathcal{L}_{\text{all}}} %
\newcommand{\pcddist}{NormalizedDistance}

\newcommand{\pcddistSH}{ND}
\newcommand{\pcddistSHspace}{ND }
\newcommand{\pcdacc}{DistanceAccu@0.2}

\newcommand{\pcdaccSH}{DAc}
\newcommand{\pcdaccSHspace}{DAc }
\newcommand{\maespace}{mAE }

\newcommand{\beginsupplement}{
    \setcounter{table}{0}
    \renewcommand{\thetable}{S\arabic{table}}%
    \setcounter{figure}{0}
    \renewcommand{\thefigure}{S\arabic{figure}}%
    \setcounter{equation}{0}
    \renewcommand{\theequation}{S\arabic{equation}}
}

\abstract{
Recent sparse multi-view scene reconstruction advances like \dst\ and \mastthreer no longer require camera calibration and camera pose estimation. However, they only process a pair of views at a time to infer pixel-aligned pointmaps. When dealing with more than two views, a combinatorial number of error prone pairwise reconstructions are usually followed by an expensive global optimization, which often fails to rectify the 
pairwise reconstruction errors.  To handle more views, reduce errors, and improve inference time, we propose the fast single-stage feed-forward network \mvdustthreer. At its core are multi-view decoder blocks which exchange information across any number of views while considering one reference view. To make our method  robust to reference view selection, we further propose \mvdustthreerp, which employs cross-reference-view blocks to fuse information across different reference view choices. To further enable novel view synthesis, we extend both by adding and jointly training Gaussian splatting heads. Experiments on multi-view stereo reconstruction, multi-view pose estimation, and novel view synthesis confirm that our methods improve significantly upon prior art. Code will be released.
}

\begin{document}

\maketitle

\begin{center}
    \includegraphics[width=1.0\columnwidth]{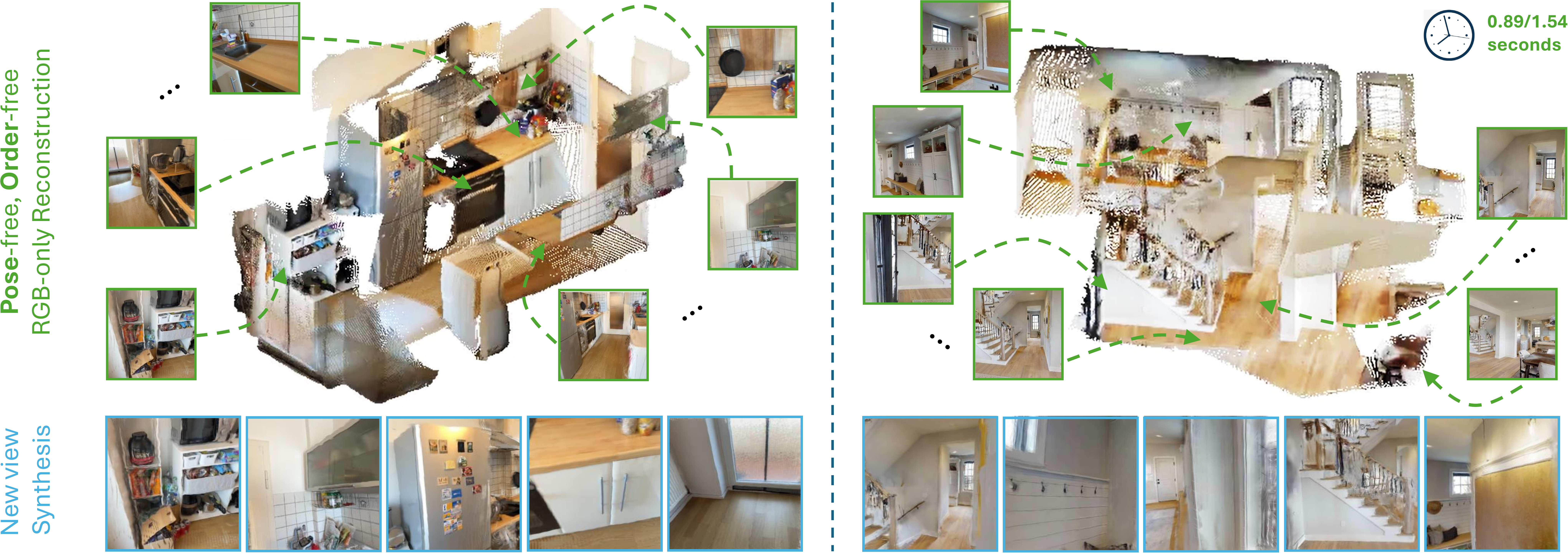}
    \captionof{figure}{The proposed Multi-View Dense Unconstrained Stereo 3D Reconstruction Prime (\mvdustthreerp) is able to reconstructs large scenes from multiple pose-free RGB views. \textbf{Top row}: one single-room scene and one large multi-room scene reconstructed by \mvdustthreerp in 0.89 and 1.54 seconds using 12 and 20 input views respectively (only a subset is shown for visualization). \textbf{Bottom row}: \mvdustthreerp is able to synthesize novel views by predicting pixel-aligned Gaussian parameters. Reconstruction of such large scenes are challenging for prior methods (\eg \dustthreer~\citep{Dust3r}). See \cref{fig:recons_results_comp} and appendix for more results with comparison.}
    \label{fig:teaser}
    
\end{center}

\section{Introduction}
\label{sec:intro}

Multi-view scene reconstruction as shown in~\cref{fig:teaser} has been a fundamental task in 3D computer vision for decades~\citep{hartley2003multiple}. It is widely applicable  in mixed reality~\citep{SceneScript}, city reconstruction~\citep{Meganerf}, autonomous driving simulation~\citep{Autosplat, Pandaset}, robotics~\citep{robotActiveVision} and archaeology~\citep{peppa2018archaeological}. Classic methods decompose multi-view scene reconstruction into sub-tasks, including camera calibration~\citep{zhang2011cameracalib, bougnoux1998projective}, pose estimation~\citep{quan1999linear, Dsac}, feature detection and matching~\citep{sift, orb, lift}, structure from motion (SfM)~\citep{schonberger2016structure}, bundle adjustment~\citep{bundleAdjustment,cadena2016past}, \etc, %
and assemble individual components into a pipeline. Recent approaches adopt a learning-based paradigm for those sub-tasks~\citep{Geoneus}, explore different neural scene representations (\eg, Neural Signed Distance Functions~\citep{Deepsdf, ma2023towards, Geoneus}, Neural Radiance Fields~\citep{Nerf, triplanenerf}, Gaussian Splatting~\citep{3dGaussian, 2dgaussian}), and build more end-to-end pipelines to reconstruct objects~\citep{Lgm, Grm} and scenes~\citep{gslrm, pixelsplat, Mvsplat}. %
While these approaches have enjoyed quite some success, 
they often require prior knowledge or nontrivial pre-processing to obtain camera parameters and poses.

More recently, novel multi-view scene reconstruction approaches, such as \dustthreer~\citep{Dust3r} and \mastthreer~\citep{MASt3R}, directly process an unordered set of unposed rgb views, \ie, camera intrinsics and poses are unknown. These methods process two views, \ie, a chosen reference view and another source view, at a time, and directly infer pixel-aligned 3D pointmaps in the reference view's camera coordinate system. %
To handle a larger set of %
input views,  a combinatorial number of %
pairwise inferences is followed by a second stage of global optimization to align the local pairwise reconstructions into a single global coordinate system. 

\begin{figure}[t]
  \centering
  \includegraphics[width=1.0\textwidth]{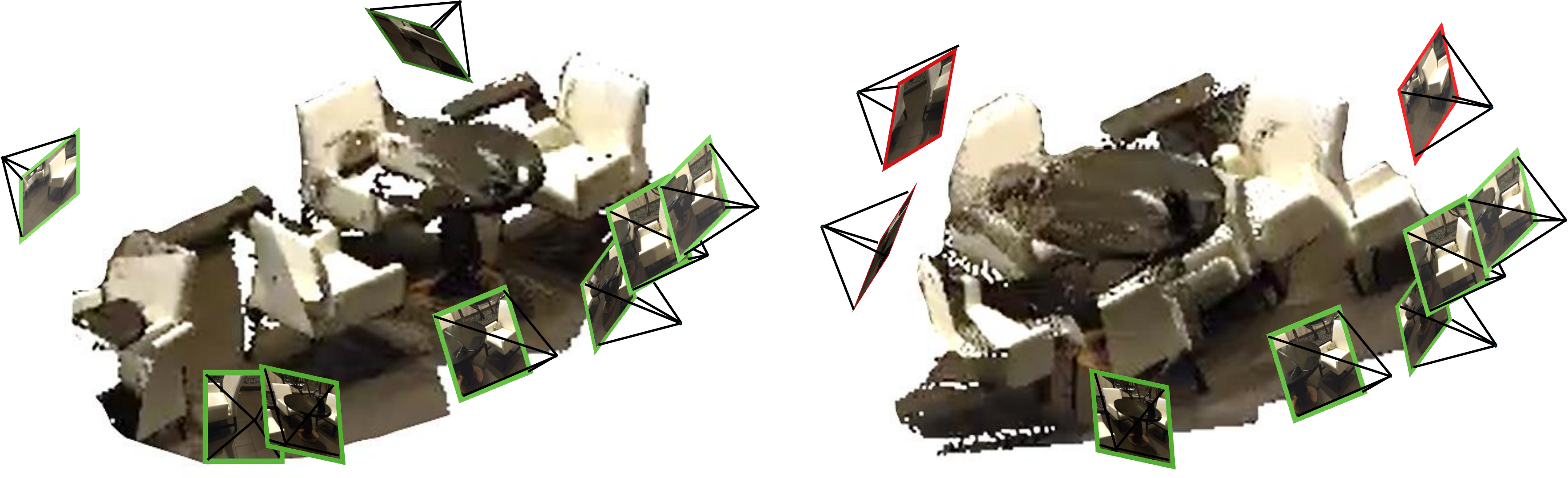}
  \caption{\textbf{Left}: Groundtruth scene with 8 views: Three chairs surrounding one table and one more chair next to another table. \textbf{Right}: reconstruction and pose estimation of \dustthreer with global optimization: all chairs incorrectly surround one table. Wrong poses are marked in \textcolor{red}{red}.
  }
  \label{fig:dstFailure}
\end{figure}

While evaluation results of these methods are promising %
on object-centric DTU data~\citep{DTUdataset}, we point out inefficiencies in reconstructing scenes, as stereo cues in 2-view input could be ambiguous. %
Further, despite plausible reconstructions of individual 2-view inputs, conflicts often arise when aligning them in a global coordinate system. Such conflicts can be challenging to  resolve by a global optimization, which only rotates  pairwise predictions but doesn't rectify wrong pairwise matches. As a consequence, scene reconstructions exhibit misaligned pointmaps. One example is shown in~\cref{fig:dstFailure}.

To address the aforementioned issues, we propose the single-stage network \textbf{Multi-View Dense Unconstrained Stereo 3D Reconstruction} (\textbf{\mvdustthreer}), which jointly processes a large number of input views in one feed-forward pass, and completely removes the cascaded global optimization used in prior arts. To achieve this, we employ multi-view decoder blocks, which jointly learn not only all pairwise relationships between a chosen reference view and all the other source views, but also appropriately address pairwise relationships among all source views. Moreover, our training recipe encourages  the predicted per-view pointmaps to adhere to the same reference camera coordinate system, which waives the need for a subsequent global optimization.

When reconstructing a large scene from sparse multi-view images, the stereo cues between the one selected reference view and certain source views could be insufficient. %
This is because significant changes in camera poses
make it difficult to directly infer the  relation between the reference view and those source views. Therefore, long-range information propagation is required for those source views. To  handle this efficiently, we further present \textbf{\mvdustthreerp}. It operates on \emph{a set of reference views} and employs \crossrefview attention blocks for  effective long-range information propagation. See \cref{fig:teaser} for its reconstructions.

On the three benchmark scene-level datasets  HM3D~\citep{hm3d}, ScanNet~\citep{scannet}, and MP3D~\citep{chang2017matterport3d}, we demonstrate that \mvdustthreer achieves significantly better results on the tasks of Multi-View Stereo (MVS) reconstruction and Multi-View Pose Estimation (MVPE) while being $48\sim78\times$ faster than \dustthreer. In addition, \mvdustthreerp is able to improve the reconstruction quality especially on harder settings, while still inferring one order of magnitude faster than \dustthreer. %

To extend both of our methods towards Novel View Synthesis (NVS), %
we further attach  lightweight prediction heads which regress to  3D Gaussian attributes. The predicted per-view Gaussian primitives are transformed into the coordinate of a target view before splatting-based rendering~\citep{3dGaussian}. %
Using this, we also show that our models outperform %
\dustthreer with heuristically designed 3D Gaussian parameters under the standard photometric evaluation protocol. The gains can be attributed to the more accurate predictions of the Gaussian locations by our method. We summarize our contributions as follows: %

\begin{itemize}[leftmargin=*]

\item We present \textbf{\mvdustthreer}, a novel feed-forward network for pose-free scene reconstruction from sparse multi-view input. %
It not only runs $48 \sim 78 \times$ faster than \dustthreer for $4\sim 24$ views, but also reduces Chamfer distance on 3 challenging evaluation datasets HM3D~\citep{hm3d}, ScanNet~\citep{scannet}, and MP3D~\citep{chang2017matterport3d} by $2.8\times$, $2\times$ and $1.6\times$ for smaller scenes of average size 2.2, 7.5, 19.3 ($m^2$) with 4-view input, and $3.2\times$, $1.9\times$ and $2.1\times$ for larger scenes of average size 3.3, 17.9, 37.3 ($m^2$) with 24-view input.

\item We present \textbf{\mvdustthreerp}, which improves \mvdustthreer by using multiple reference views, %
addressing the challenges which occur when inferring  relations between all input views via a single  reference view. We validate, \mvdustthreerp performs well across all tasks, number of views, and on all three datasets. For example, for MVS reconstruction, it further reduces Chamfer distance on 3 datasets by $2.6\times,1.6\times,1.8\times$ 
for large scenes with 24-view input, while still running $14\times$ faster than \dustthreer. 

\item We extend both  networks to support NVS by adding Gaussian splatting heads to predict per-pixel Gaussian attributes.
With joint training of all layers using both reconstruction loss and view rendering loss, we demonstrate that the model outperforms a \dustthreer-based baseline significantly.

\end{itemize}

\section{Related Work}
\label{sec:rw}

\noindent \textbf{Structure-from-Motion (SfM).} SfM methods reconstruct sparse scene geometry from a set of images and estimate individual camera poses. %
For this, SfM is often addressed in a few independent steps,
including detecting/describing/matching local features across multiple views (\eg, SIFT~\citep{sift}, ORB~\citep{orb}, LIFT~\citep{lift}), triangulating features to estimate sparse 3D geometry and camera poses (\eg, COLMAP~\citep{schonberger2016structure}), applying bundle adjustment over many views (see \citet{bundleAdjustment} for an overview), \etc. Though steady progress has been made in the past decades~\citep{ozyecsil2017survey}, and a large number of applications have been enabled~\citep{westoby2012structure, iglhaut2019structure, carrivick2016structure}, the classic SfM pipeline solves sub-tasks individually and sequentially, accumulating errors. More recent SfM methods improve traditional pipeline with learnable components~\citep{detectorFreeSFM, VGGSFM}. MASt3R-SfM~\citep{mast3rsfm} extends MASt3R~\citep{MASt3R}, which only produces local reconstructions for 2-view input, to perform global optimization for aligning local reconstructions via gradient descent to minimize 3D matching loss.

\noindent \textbf{Multi-View Stereo.} MVS reconstructs dense 3D scene geometry from multiple views~\citep{mvs}, often in the form of 3D points. In the classic PatchMatch-based framework~\citep{zheng2014patchmatch}, per-pixel depth in the reference image is estimated from a set of unstructured source images via patch matching under a homography transform~\citep{schonberger2016pixelwise}. Subsequent work has substantially improved feature matching~\citep{wang2023adaptive, zhou2018learning, Patchmatchnet} and depth estimation~\citep{galliani2015massively, R3d3, xu2019multi}. More recent learning-based  approaches~\citep{Mvsnet, Mvdepthnet} often build an end-to-end pipeline, where deep models extract visual features, model cross-view correspondences (\eg, cost volume~\citep{gu2020cascade}), and regress depth maps~\citep{yao2019recurrent}. Note, with few exceptions~\citep{yariv2020multiview}, most approaches require prior knowledge of camera intrinsics from SfM or camera calibration. Our \mvdustthreer network also processes sparse multi-view input, but does not require prior knowledge of camera parameters. %

\noindent \textbf{Neural Scene Reconstruction.} Compared to classic methods, which reconstructs a scene using either explicit representations (\eg, 3D point, mesh) or implicit representations (\eg, signed distance function~\citep{Kinectfusion}), recent approaches adopt different neural representations~\citep{sitzmann2019scene}, including Neural Distance Fields~\citep{chibane2020neural, Dist},
Neural Radiance Fields (NeRFs)~\citep{Nerf, mipnerf, Meganerf, panopticnerf, nopenerf}, Gaussian Splatting~\citep{3dGaussian, 2dgaussian, mipsplatting}, and their combination~\citep{triplanegaussian}. Many of them require slow per-scene optimization to attain accurate results, while more recent methods explore the use of feed-forward networks for generalizable reconstruction at a fraction of the time, including those for generating Neural Distance Functions~\citep{Deepsdf, Metasdf, chibane2020neural}, NeRFs~\citep{pixelnerf, AttnRend, Mvsnerf}, and Gaussians~\citep{latsplat, Splatterimage, pixelsplat, Mvsplat}. Note, neural scene reconstruction  often requires input views with known camera poses, albeit quite a few exceptions exist, such as CoPoNeRF~\citep{CoPoNeRF}, Splatt3R~\citep{splatt3r}, and NoPoSplat~\citep{nopo}. For example, NoPoSplat predicts %
3D Gaussians in the same camera coordinates, akin to the key idea of \dustthreer. 
However, those pose-free methods  primarily focus on inference with 2 input views. It is not clear how they  perform when processing sparse multi-view input. In contrast, our models \mvdustthreer, \mvdustthreerp equipped with 3D Gaussian splatting heads, not only waive the need for camera pose, but also  reconstruct large scenes from multiple views in a single feed-forward pass.

\noindent \textbf{Dense Unconstrained Scene Reconstructions from Multi-View Input.} %
To bypass estimation of camera parameters and poses,  recent works like \dustthreer~\citep{Dust3r} and \mastthreer~\citep{MASt3R} propose a new approach:  directly regress pixel-aligned 3D pointmaps for pairs of input views. %
An expensive $2^\text{nd}$ stage global optimization is required to align all pairwise reconstructions in the same coordinate system. Both \dustthreer and \mastthreer are only evaluated on object-centric  DTU data~\citep{DTUdataset} where all views are concentrated in a small region. Notably, methods are not validated if their 2-stage pipeline  excels at reconstructing larger scenes captured with sparse multi-view input. Subsequently,  Spann3R~\citep{Spann3R} augments \dustthreer with a spatial memory to process an ordered set of images. Although capable of performing online scene reconstruction for object-centric scenes, for larger scenes, Spann3R is more likely to drift, generating a misaligned reconstruction due to the limited size of the spatial memory and the lack of globally aligning reconstructions. %
In contrast, our \mvdustthreerp performs offline scene reconstruction by processing all input views (up to 24 in our experiments) at once. Different from \dustthreer, it does not require global optimization because the predicted per-view pointmaps are already globally aligned.

\noindent \textbf{Generative models for 3D reconstruction.} Reconstructing scenes from a small number of views is challenging, in particular for unseen areas. Recent advances such as InFusion~\citep{InFusion}, ZeroNVS~\citep{Zeronvs}, Reconfusion~\citep{Reconfusion}, and ReconX~\citep{Reconx}  exploit priors encoded in image and video generative models~\citep{blattmann2023align, stableVideoDiffusion, SongCXKTYY23}. We leave benefit from  image priors of  diffusion models as future work. %

\section{Method}
\label{sec:method}

Our goal is to densely reconstruct a scene given a sparse set of  rgb images  with unknown camera intrinsics and poses. Following  \dustthreer, our model predicts 3D pointmaps aligned with 2D pixels for each view. %
Different from \dustthreer, our model jointly predicts 3D pointmaps for any number of input views in a single forward pass. 
Formally, given $N$ input image views of a scene $\{I^{v}\}_{v=1}^{N}$, where $I^{v} \in \mathbb{R}^{H \times W\times 3}$, from which we select one reference $r\in\{1, \dots, N\}$, our goal is to predict per-view 3D pointmaps $\{X^{v,r}\}_{v=1}^{N}$. Note, the 3D pointmap $X^{v,r} \in \mathbb{R}^{H \times W\times 3}$ denotes the coordinates of 3D points for image $I^v$ in the camera coordinate system of the reference view $r$. 

In~\cref{sec:mvdust3r}, we introduce our Multi-View Dense Unconstrained Stereo 3D Reconstruction (\textbf{\mvdustthreer}) network to efficiently processes all input views in one pass and without subsequent global optimization, while considering a single chosen reference view. 
In~\cref{sec:mvdust3rp}, we present \textbf{\mvdustthreerp}, which processes all input views while considering multiple reference views. 
Finally, to support novel view synthesis, in~\cref{sec:mvdust3rpp}, %
we augment our networks  with Gaussian heads to predict pixel-aligned 3D Gaussians.

\begin{figure}[t]
\centering
  \includegraphics[width=0.7\textwidth]{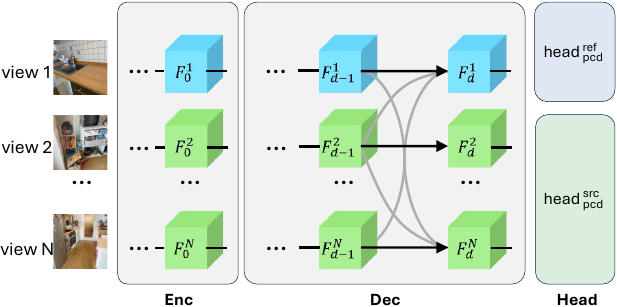}
  \vspace{-0.1cm}
  \caption{Overview of \mvdustthreer. 
  Visual tokens for the reference view and other source views are shown in \textcolor{blue}{Blue} and \textcolor{green}{Green}. %
  Black straight solid lines indicate the primary token flow while gray lines indicate  secondary token flow.
  }
  \label{fig:mvdust3r_model}
\vspace{-0.5cm}
\end{figure}

\subsection{\mvdustthreer}
\label{sec:mvdust3r}

\noindent\textbf{A Multi-View Model Architecture.} 
As shown in~\cref{fig:mvdust3r_model}, \mvdustthreer consists of an encoder to transform images into visual tokens, decoder blocks to fuse tokens across views, and regression heads to predict per-view 3D pointmaps aligned with 2D pixels. Different from \dustthreer, our network %
uses decoder blocks to fuse tokens across \emph{all}  views rather than independently fusing only tokens for two views at a time. %
Concretely, a ViT~\citep{visionTransformer} encoder with shared weights, denoted as \texttt{Enc}, is first applied on input views $\{I^{v}\}_{v=1}^{N}$ to compute initial visual tokens $\{F^v_0\}_{v=1}^N$, \ie, $F^v_0 = \texttt{Enc}(I^v)$.
Note, the resolution of the encoder output features is $16\times$ smaller than the input image before being flattened into a sequence of tokens. 

To fuse the tokens, two types of decoders %
are used, one for the chosen reference view and one for the remaining source views. They share the same architecture but their weights differ. Each decoder consists of $D$ decoder blocks referred to as $\texttt{DecBlock}_d^\text{ref}$ and $\texttt{DecBlock}_d^\text{src}$ for $d\in\{1, \dots, D\}$. Their difference is, ${\texttt{DecBlock}}^{\text{ref}}_d$ is dedicated to update reference view tokens $F^r$, while ${\texttt{DecBlock}}^{\text{src}}_d$ updates tokens $\{F^v\}_{v \neq r}$ from \textit{all other} source views. 
Each decoder block takes as input a set of primary tokens from one view, and a set of secondary tokens from other views. In each block, a self-attention layer is applied to primary tokens only, %
and a cross-attention layer fuses primary tokens with secondary tokens before a final MLP is applied on the primary tokens. Layer norm is also applied before both attentions and the MLP. %
Using those, the decoder computes the final token representations $F^v_D$  via 
\begin{equation}
    F^v_d = \begin{cases}
        \texttt{DecBlock}^\text{ref}_d(F^v_{d-1},{\cal F}^{-v}_{d-1}) ~\text{if}~ v=r,\\
        \texttt{DecBlock}^\text{src}_d(F^v_{d-1},{\cal F}^{-v}_{d-1}) ~\text{otherwise}. 
    \end{cases}
\vspace{-0.1cm}
\end{equation}
Here, the secondary tokens $\mathcal{F}^{-v}_{d} = \{F^{1}_{d},\dots,F^{v-1}_{d},F^{v+1}_{d}, \dots ,F^{N}_{d}\}$ subsume tokens from all views other than the view of the primary tokens $F^v_d$.

To finally predict the per-view 3D pointmaps, %
we use two heads: $\texttt{Head}^{\text{ref}}_\text{pcd}$ for the reference view  and  $\texttt{Head}^{\text{src}}_\text{pcd}$ for all other views. They share the same architecture but use different weights. Each consists of a linear projection layer and a pixel shuffle layer with an upscale factor of $16$ to restore the original input image resolution. As in \dustthreer, the head predicts 3D pointmaps %
$X^{v,r} \in \mathbb{R}^{H\times W \times 3}$ 
and confidence maps $C^{v,r} \in \mathbb{R}^{H\times W}$ via %
\begin{equation}
    X^{v,r}, C^{v,r} = \begin{cases}
        \texttt{Head}^{\text{ref}}_\text{pcd}(F^v_D) ~\text{if}~ v = r,\\
        \texttt{Head}^{\text{src}}_\text{pcd}(F^v_D) ~\text{otherwise}.
    \end{cases}
\label{eqn:head}
\end{equation}
Note that \dustthreer  is a special case of \mvdustthreer if the number of views $N=2$. However, for multiple input views, \mvdustthreer will update primary tokens using a much larger set of secondary tokens. Hence, it is able to benefit from many more views. %
Importantly, as our architecture components and structure only differ slightly from those in \dustthreer (additional skip connection and conv net), we have only marginally more trainable parameters. Since, the  number of parameters in \mvdustthreer is almost identical to  \dustthreer, \mvdustthreer can beneficially be initialized using pre-trained  \dustthreer  weights. %

\noindent\textbf{Training Recipe.} 
Inspired by \dustthreer, we use a confidence-aware pointmap regression loss $\lconf$, \ie, 
\begin{subequations}
\begin{equation}
\lconf=\sum_{v\in\{1, \dots, N\}}\sum_{p\in{P^v}}C_p^{v,r}\ell_{\text{regr}}(v,p) - \beta \log C^{v,r}_p, \\
\label{eqn:confReconLoss}
\end{equation} 
\begin{equation}
\text{where} \quad\ell_\text{regr}(v,p)=\left\|\frac{1}{z}X_p^{v,r}-\frac{1}{\bar{z}}\bar{X}_p^{v,r}\right\|.
\end{equation} 
\end{subequations}
Here, $P_v$ denotes the set of valid pixels in view $v$ where groundtruth 3D points are well defined. $\beta$ controls the weight of the regularization term. %
The pointmap regression loss $\ell_{\text{regr}}$ measures the difference between predicted and groundtruth 3D points after normalization, which is needed to resolve the scale ambiguity between prediction and groundtruth.
It uses $\bar{X}_p^{v,r}$, the groundtruth 3D point of pixel $p$ of view $v$ in the reference view $r$. The scale normalization factor $z=\text{norm}(\mathcal{X}^{\{v\},r})$ and $\bar{z}=\text{norm}(\bar{\mathcal{X}}^{\{v\},r})$ are computed as the average distance of valid 3D points to the coordinate origin in all views, for prediction and groundtruth, respectively.

\subsection{\mvdustthreerp}
\label{sec:mvdust3rp}

\begin{figure}[t]
  \centering
  \includegraphics[width=0.9\textwidth]{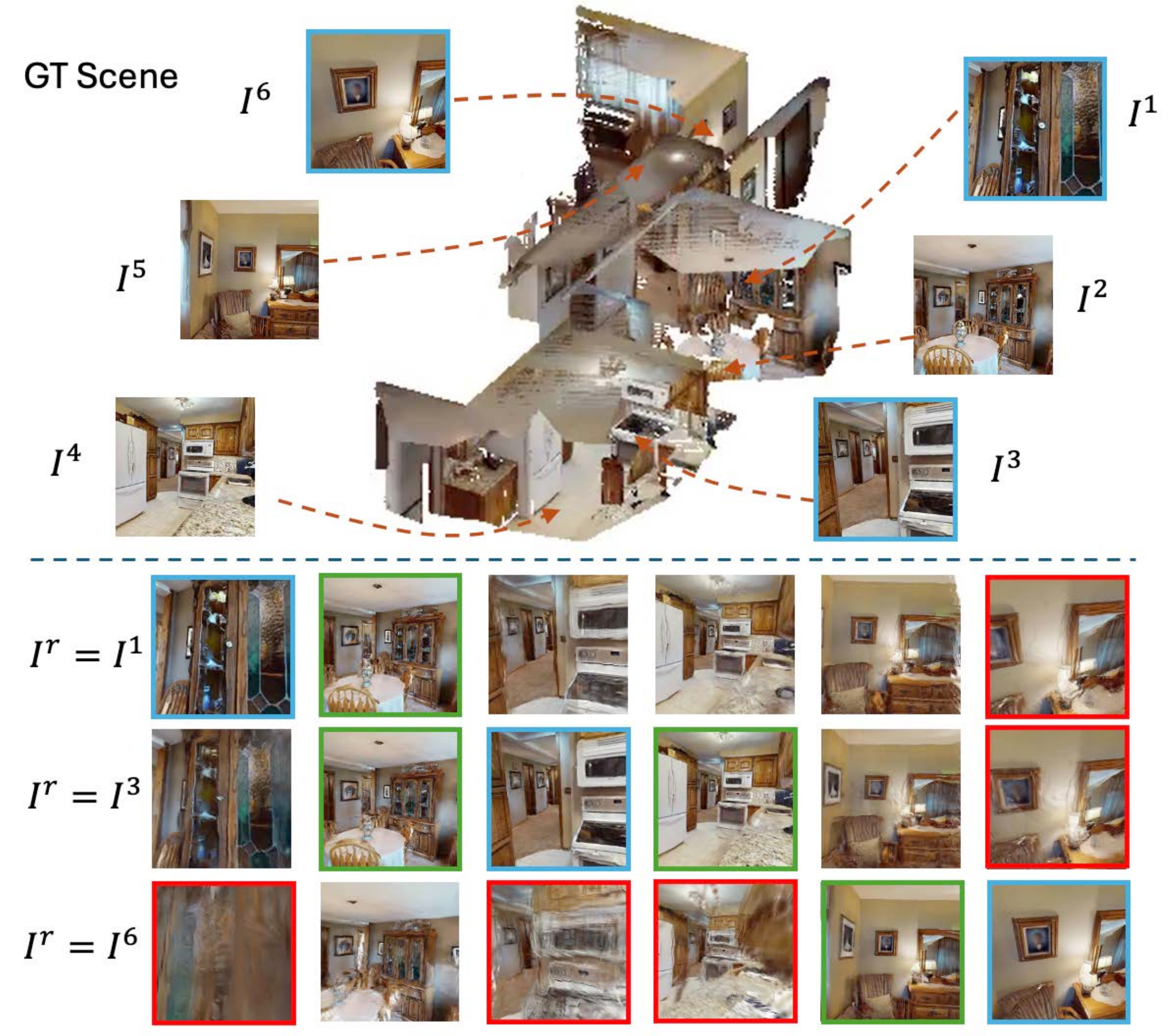}
  \vspace{-0.2cm}
  \caption{\textbf{Top}: A multi-room scene: 16  views are sampled as input to \mvdustthreer. For clarity, only 6 are shown. 3 of them are reference view candidates, highlighted in \textcolor{blue}{blue}. \textbf{Bottom}: In each row, we select a different reference view and  render the reconstructed scene from 6 input views. Renderings in good and poor quality are highlighed in \textcolor{green}{green} and \textcolor{red}{red}.
  As the viewpoint change between the input view and the reference view increases,  quality of the reconstructed scene geometry in that input view  decreases.}
  \label{fig:mref}
\end{figure}

As shown in~\cref{fig:mref}, for different reference view choices, the quality of the scene reconstructed by \mvdustthreer varies spatially. The predicted pointmap for an input source view tends to be better when the viewpoint change to the reference view is small, and deteriorates as the viewpoint change increases. %
However, to reconstruct a large scene with a sparse set of input views, a single reference view with only moderate viewpoint changes to all other source views is unlikely to exist. Therefore, it is difficult to reconstruct scene geometry equally well everywhere with a single selected reference view.
To address this, we propose \textbf{\mvdustthreerp}, which selects multiple views as the reference view, and jointly predicts pointmaps for all input views in the camera coordinate of each selected reference view. We hypothesize: while pointmaps of certain input views are difficult to predict for one reference view, they are easier to predict for a different reference view (\eg, smaller viewpoint change, more salient matching patterns). To holistically improve the pointmap prediction of all input views, we include a novel \crossrefview block into \mvdustthreerp.

\begin{figure}[t]
  \centering
  \includegraphics[width=0.7\textwidth]{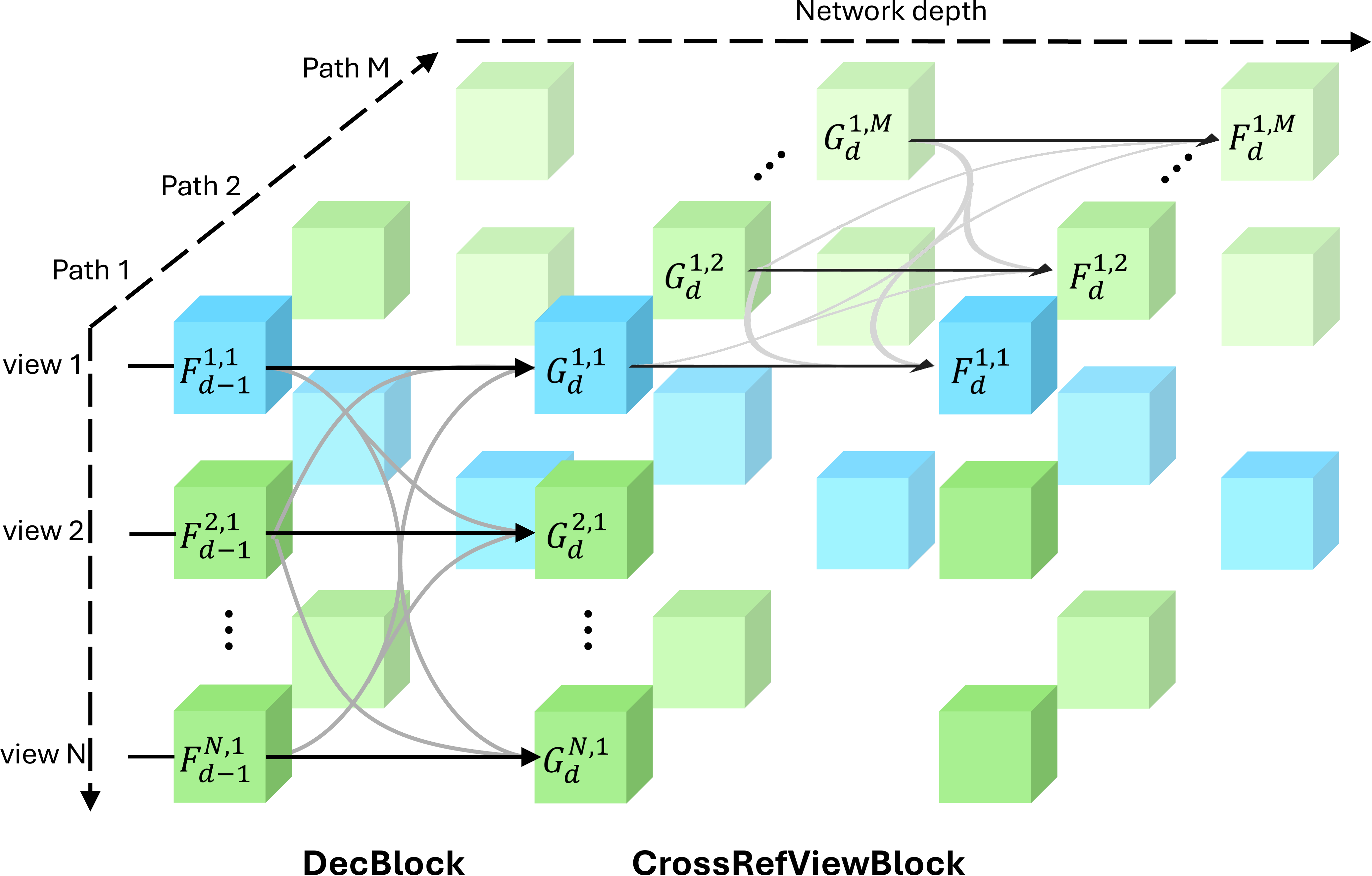}
  \caption{\texttt{DecBlock} and \texttt{CrossRefViewBlock} in \mvdustthreerp: tokens of the reference and other views are highlighted in \textcolor{blue}{blue} and \textcolor{green}{green}, respectively. Each model path uses a different reference view. 
  For clarity, only 1 of stacked \texttt{DecBlock} and \texttt{CrossRefViewBlock} are shown.
  }
  \label{fig:mpath}
\end{figure}

\noindent\textbf{A Multi-Path Model Architecture.} 
Let $R=\{r^m\}_{m=1}^M$ denote a set of $M$ reference views randomly chosen from an unordered set of input views. We adopt the same decoder blocks from \mvdustthreer, deploy them in a multi-path model architecture (see \cref{fig:mpath}), and use them to compute a reference-view dependent intermediate representation $G_d^{v, m}$ at decoder layer $d$ for input view $v$ \textit{and reference view $r^m$}:
\begin{numcases}{G_d^{v, m}=}
        \texttt{DecBlock}^\text{ref}_d(F_{d-1}^{v, m}, {\cal F}_{d-1}^{-v, m}) ~\text{if}~ v=r^m,
        \\
        \texttt{DecBlock}^\text{src}_d(F_{d-1}^{v, m}, {\cal F}_{d-1}^{-v, m})~\text{otherwise},
\end{numcases}

\begin{equation}
F_{d}^{v, m}=\texttt{CrossRefViewBlock}_d(G_d^{v, m}, \mathcal{G}_d^{v, -m}).
\label{eqn:crossPathTokenUpdate}
\end{equation} 
Here, $\mathcal{F}_{d-1}^{-v, m} = \{F_{d-1}^{1, m},\dots,F_{d-1}^{v-1, m},F_{d-1}^{v+1, m}, \dots ,F_{d-1}^{N, m}\}$. 
As shown in~\cref{fig:mpath}, we fuse and update per-view tokens computed under different reference views by adding a \textit{\crossrefview} block after each decoder block (\cref{eqn:crossPathTokenUpdate}), where $\mathcal{G}_d^{v, -m} = \{G^{v, 1}_{d},\dots, G^{v,m-1}_d, G^{v,m+1}_d, \dots,G^{v, M}_d\}$.
Following~\cref{eqn:head}, we compute per-view pointmaps $X^{v, m}$ and confidence map $C^{v, m}$ under each reference view $r^m$. 

\noindent\textbf{Training Recipe.} 
Compared with \mvdustthreer, \mvdustthreerp only adds a small number of additonal trainable parameters via the \crossrefview blocks (see appendix). %
During training, a random subset of $M$ input views are selected as the reference views. We average the pointmap regression losses  in~\cref{eqn:confReconLoss} for all reference views. %

\noindent\textbf{Model Inference.} At inference time, we uniformly select a subset of $M$ input views as the reference views, while the 1st input view is always selected. A model with $M$ paths is used but the final per-view pointmap predictions are computed using the heads in the 1st path.

\subsection{MV-DUSt3R(+) for Novel View Synthesis}
\label{sec:mvdust3rpp}
Next we extend our networks to support NVS with Gaussian primitives~\citep{3dGaussian}. For clarity, below we use \mvdustthreerp as an example. \mvdustthreer can be extended similarly.

\noindent\textbf{Gaussian Head.} 
We add a separate set of heads to predict per-pixel Gaussian parameters, including scaling factor $S^{v,m}\in \mathbb{R}^{H\times W \times 3}$, rotation quaternion $q^{v,m}\in \mathbb{R}^{H\times W \times 4}$, and opacity $\alpha^{v,m}\in \mathbb{R}^{H\times W}$. We add Gaussian heads $\texttt{Head}^{\text{ref}}_\text{3DGS}$ and $\texttt{Head}^{\text{src}}_\text{3DGS}$ for reference and other views: 
\begin{numcases}{ S^{v,m},q^{v,m},\alpha^{v,m}=}
        \texttt{Head}^\text{ref}_\text{3DGS}(F^{v,m}_D) ~\text{if}~ v=r^m,
        \\
        \texttt{Head}^\text{src}_\text{3DGS}(F^{v,m}_D) ~\text{otherwise}.
\end{numcases}
For other Gaussian parameters, we use the predicted pointmap $X^{v,m}$ as the center, the pixel color $I^{v}$ as the color and fix the spherical harmonics degree to be 0.

\noindent\textbf{Training Recipe.}
During training, for a chosen reference view $r^m$, we perform differentiable splatting-based rendering~\citep{3dGaussian} to generate rendering predictions for both input views and novel views. Following prior approaches~\citep{Splatterimage, pixelsplat, Mvsplat}, we use a weighted sum of $L^2$ pixel difference loss and perceptual similarity loss LPIPS as the rendering loss $\lrender$ to train the Gaussian heads. The final training loss includes both $\lconf$ and $\lrender$ (for details see appendix).

\section{Experiments}
\label{sec:exp}

\subsection{Datasets}

Our training data includes ScanNet~\citep{scannet}, ScanNet++~\citep{scannet++}, HM3D~\citep{hm3d}, and Gibson~\citep{gibson}. 
Note, all of them are also used by \dustthreer.  For evaluation, we use datasets MP3D~\citep{chang2017matterport3d}, HM3D~\citep{hm3d}, and ScanNet~\citep{scannet}. While ScanNet scenes are often small single-room  sized and with low diversity, scenes in MP3D and HM3D are often large multi-room sized and with high diversity. MP3D also contains outdoor scenes. See~\cref{tab:evalDatasets} to compare evaluation datasets. We use the same train/test split as  \dustthreer, and our training data is a subset of \dustthreer's training data (for details see appendix).


\begin{table}[t]
    \centering
    \resizebox{0.5\columnwidth}{!}{
    \setlength{\tabcolsep}{2pt}
        \begin{tabular}{c|c|c}
        Dataset & Eval setting & Scene type\\
        \specialrule{.2em}{.1em}{.1em}
        HM3D & Supervised &multi-room (large)\\
        \scannet & Supervised & single-room (small)\\
        MP3D &  Zero-shot & multi-room \& outdoor (largest)\\
            \bottomrule 
        \end{tabular}
    }
    \caption{\textbf{Evaluation datasets comparisons}. 
    }
    \label{tab:evalDatasets}
\end{table}

\noindent\textbf{Trajectory Generation.} 
To generate a set of input views $\{I^v\}_{v=1}^N$ for $N>2$, we first randomly select one frame and initialize the current scene point cloud using its data. Then we sequentially sample more candidate frames. We retain a candidate frame and add its corresponding point cloud to the current scene, if the overlap between the candidate frame's point cloud and the current scene point cloud is between a lower threshold $t_\text{min}$ and an upper bound $t_\text{max}$. 

\noindent\textbf{Training Trajectories.} 
To sample the training set trajectories, we employ two choices of thresholds: $\text{(}t_{\text{min}}, t_{\text{max}} \text{)} \in \{\text{(}30\%, 70\%\text{)}, \text{(}30\%, 100\%\text{)} \} $. From ScanNet and ScanNet++, we sample 1K trajectories of $10$ views per scene, and a total of 3.2M trajectories. On HM3D and Gibson, where the scene is often larger, we sample 6K trajectories per scene with 10 views each, and a total of 7.8M trajectories. 

\noindent\textbf{Test Trajectories.} 
For the test set, we generate $1K$ trajectories per dataset. %
To support  evaluation with a larger number of inputs views, we sample $30$ views per trajectory.

\subsection{Implementation Details}
We process input views at resolution $224\times 224$. We utilize $64$ Nvidia H100 GPUs for the model training. %
To initialize, 
\dustthreer model weights are used. %
We use the first $N=8$ views of each trajectory as input views, and randomly select 1 view as the reference view for \mvdustthreer and $M=4$ views for \mvdustthreerp. %
We train for 100 epochs using 150K trajectories per epoch, which takes 180 hours. For MVS reconstruction evaluation, to assess the performance of each method in reconstructing scenes of variable sizes, we report results with input views ranging from 4 to 24 views. For NVS evaluation, we use the remaining 6 views as novel views. Below we report  results on all evaluation datasets for all choices of the number of input views using only one \mvdustthreer model and one \mvdustthreerp model.

\begin{table}[t]
    \centering
    \resizebox{0.9\columnwidth}{!}{
    \setlength{\tabcolsep}{2pt}
        \begin{tabular}{c|c|c|ccc|ccc|ccc|c}
            & \multirow{2}{*}{Method}& \multirow{2}{*}{GO} & \multicolumn{3}{c|}{HM3D} & \multicolumn{3}{c|}{ScanNet} & \multicolumn{3}{c}{MP3D} & Time\\\cline{4-12}
            &  & & ND $\downarrow$ & DAc $\uparrow$ & CD $\downarrow$ & ND $\downarrow$ & DAc $\uparrow$ & CD $\downarrow$ & ND $\downarrow$ & DAc $\uparrow$ & CD $\downarrow$ & (sec) \\
            \specialrule{.2em}{.1em}{.1em}
            \multirow{5}{*}{\rotatebox[origin=c]{90}{4 views}}
            & Spann3R & $\times$ & 37.1 & 0.0 & 225(184) & 8.9 & 19.5 & 54.7(50.1) & 42.7 & 0.0 & 248(202) & 0.36\\
            & \dustthreer & \checkmark & 1.9 & 75.1 & 5.6(2.3) & 1.3 & 89.8 & 4.0(0.4) & 3.9 & 41.7 & 40.0(5.3) & 2.42\\
            & \mvdustthreer & $\times$ & 1.1 & 92.2 & 2.0(1.1) & 1.0 & 93.3 & 2.0(0.4) & 2.5 & 62.4 & 25.3(4.1)& \cellcolor[rgb]{0.999, 0.8, 0.8}0.05\\
            & \mvdustthreerp & $\times$ & \cellcolor[rgb]{0.999, 0.8, 0.8} 1.0 & \cellcolor[rgb]{0.999, 0.8, 0.8}95.2 & \cellcolor[rgb]{0.999, 0.8, 0.8}1.5(0.9) & \cellcolor[rgb]{0.999, 0.8, 0.8}0.8 & \cellcolor[rgb]{0.999, 0.8, 0.8}94.9 & \cellcolor[rgb]{0.999, 0.8, 0.8}1.5(0.3) & \cellcolor[rgb]{0.999, 0.8, 0.8}2.2 & \cellcolor[rgb]{0.999, 0.8, 0.8}68.0 & \cellcolor[rgb]{0.999, 0.8, 0.8}19.9(3.4) & 0.29\\\cline{2-13}
            & {\color{lightgray}$\text{\mvdustthreer}_\text{oracle}$} & $\times$ & \lgr{1.0} & \lgr{94.6} & \lgr{1.5(0.7)} & \lgr{0.8} & \lgr{95.5} & \lgr{1.3(0.3)} & \lgr{2.3} & \lgr{66.6} & \lgr{20.7(4.0)}& -\\
            & {\color{lightgray}$\text{\mvdustthreerp}_\text{oracle}$} & $\times$ & \lgr{0.9} & \lgr{96.5} & \lgr{1.4(0.7)} & \lgr{0.7} & \lgr{95.8} & \lgr{1.2(0.2)} & \lgr{2.1} & \lgr{70.6} & \lgr{17.9(3.3)}& -\\\midrule 
            \multirow{5}{*}{\rotatebox[origin=c]{90}{12 views}}
            & Spann3R & $\times$ & 32.6 &  0.0 & 125(113) & 9.1 & 16.3 & 36.6(31.2) & 35.0 & 0.0 & 138(112) & 1.34\\
            & \dustthreer & \checkmark &3.9 & 30.7 & 18.1(3.4) & 1.9 & 82.6 & 4.1(0.6) & 6.6 & 12.0 & 49.6(8.3) & 8.28\\
            & \mvdustthreer & $\times$ & 1.6 & 79.5 & 3.0(1.2) & 1.4 & 86.8 & 2.3(0.8) & 3.4 & 41.3 & 22.6(5.5) & \cellcolor[rgb]{0.999, 0.8, 0.8}0.15\\
            & \mvdustthreerp & $\times$ & \cellcolor[rgb]{0.999, 0.8, 0.8}{1.2} & \cellcolor[rgb]{0.999, 0.8, 0.8}{91.5} & \cellcolor[rgb]{0.999, 0.8, 0.8}{1.8(0.7)} &\cellcolor[rgb]{0.999, 0.8, 0.8}{ 1.2} & \cellcolor[rgb]{0.999, 0.8, 0.8}{88.4} & \cellcolor[rgb]{0.999, 0.8, 0.8}{1.8(0.7)} & \cellcolor[rgb]{0.999, 0.8, 0.8}{2.6} & \cellcolor[rgb]{0.999, 0.8, 0.8}{55.0} & \cellcolor[rgb]{0.999, 0.8, 0.8}{15.1(3.8)} & 0.89\\\cline{2-13}
            & {\color{lightgray}$\text{\mvdustthreer}_\text{oracle}$} & $\times$ & \lgr{1.3} & \lgr{88.8} & \lgr{1.8(0.9)} & \lgr{1.0} & \lgr{90.6} & \lgr{1.3(0.7)} & \lgr{2.9} & \lgr{51.3} & \lgr{16.4(4.0)} & -\\
            & {\color{lightgray}$\text{\mvdustthreerp}_\text{oracle}$} & $\times$ & \lgr{1.1} & \lgr{94.8} & \lgr{1.4(0.7)} & \lgr{1.0} & \lgr{90.9} & \lgr{1.3(0.5)} & \lgr{2.5} & \lgr{59.8} & \lgr{13.6(3.5)}& -\\\midrule 
            \multirow{5}{*}{\rotatebox[origin=c]{90}{24 views}}
            & Spann3R & $\times$ & 41.7 & 0.0 & 139(121) & 11.4 & 1.6 & 37.4(35.5) & 46.6 & 0.0 & 151(121) & 2.73\\
            & \dustthreer & \checkmark & 6.8 & 7.3 & 32.4(5.2) & 2.4 & 72.6 & 5.1(1.0) & 11.4 & 2.5 & 80.9(14.3) & 27.21\\
            & \mvdustthreer & $\times$ & 3.4 & 36.7 & 10.0(3.5) & 2.2 & 75.2 & 2.7(0.9) & 6.3 & 12.2 & 38.6(13.9) & \cellcolor[rgb]{0.999, 0.8, 0.8}0.35\\
            & \mvdustthreerp & $\times$ & \cellcolor[rgb]{0.999, 0.8, 0.8}{2.1} & \cellcolor[rgb]{0.999, 0.8, 0.8}{64.5} & \cellcolor[rgb]{0.999, 0.8, 0.8}{3.9(2.0)} & \cellcolor[rgb]{0.999, 0.8, 0.8}{1.6 }& \cellcolor[rgb]{0.999, 0.8, 0.8}{81.2} & \cellcolor[rgb]{0.999, 0.8, 0.8}{1.7(0.7)} & \cellcolor[rgb]{0.999, 0.8, 0.8}{4.3} & \cellcolor[rgb]{0.999, 0.8, 0.8}{26.7} & \cellcolor[rgb]{0.999, 0.8, 0.8}{22.0(5.9)} & 1.97\\\cline{2-13}
            & {\color{lightgray}$\text{\mvdustthreer}_\text{oracle}$} & $\times$ & \lgr{2.1} & \lgr{58.9} & \lgr{3.5(2.1)} & \lgr{1.4} & \lgr{82.9} & \lgr{1.4(0.7)} & \lgr{4.4} & \lgr{22.0} & \lgr{19.9(5.1)}& -\\
            & {\color{lightgray}$\text{\mvdustthreerp}_\text{oracle}$} & $\times$ & \lgr{1.8} & \lgr{77.9} & \lgr{2.6(1.3)} & \lgr{1.3} & \lgr{85.1} & \lgr{1.3(0.6)} & \lgr{3.6} & \lgr{33.1} & \lgr{15.1(4.4)}& -\\\bottomrule 
            
        \end{tabular}
    }
    \caption{\textbf{MVS reconstruction results}. Best results are marked in \textcolor[rgb]{0.999, 0.6, 0.6}{red}.  For clarity, metric \pcddistSH\ is scaled up by $10\times$, \pcdaccSH\ is in percent $\%$, and CD is scaled  by $100\times$ with its median given in parentheses. Results of our \textbf{oracle} methods are obtained by manually selecting the best single reference view  from reference view candidates to report a possibly achievable performance. In the rightmost column, each method's time for scene reconstruction %
    is reported.}
    \label{tab:MVSReconstruction}
\end{table}

\subsection{Multi-View Stereo Reconstruction}
\noindent\textbf{Metrics.} We report Chamfer Distance (CD), as in prior work~\citep{DTUdataset, Dust3r}, as well as 2 additional metrics. \textbf{\pcddist~(\pcddistSH)}: $\ell_\text{regr}$ with zero-centering to make it scale and translation invariant; \textbf{\pcdacc~(\pcdaccSH)}:  proportion of pixels where the corresponding normalized distance between prediction and groundtruth in pointmap is $\leq 0.2$. 

\noindent\textbf{Baselines}. We compare with baselines that  reconstruct scenes from input rgb views without knowing camera intrinsics and poses. We evaluate the \dustthreer model trained at input resolution $224\times 224$ with Global Optimization (GO), and also the Spann3R~\citep{Spann3R} model on Github.

\begin{figure*}[t]
  \centering
  \includegraphics[width=1.0\textwidth]{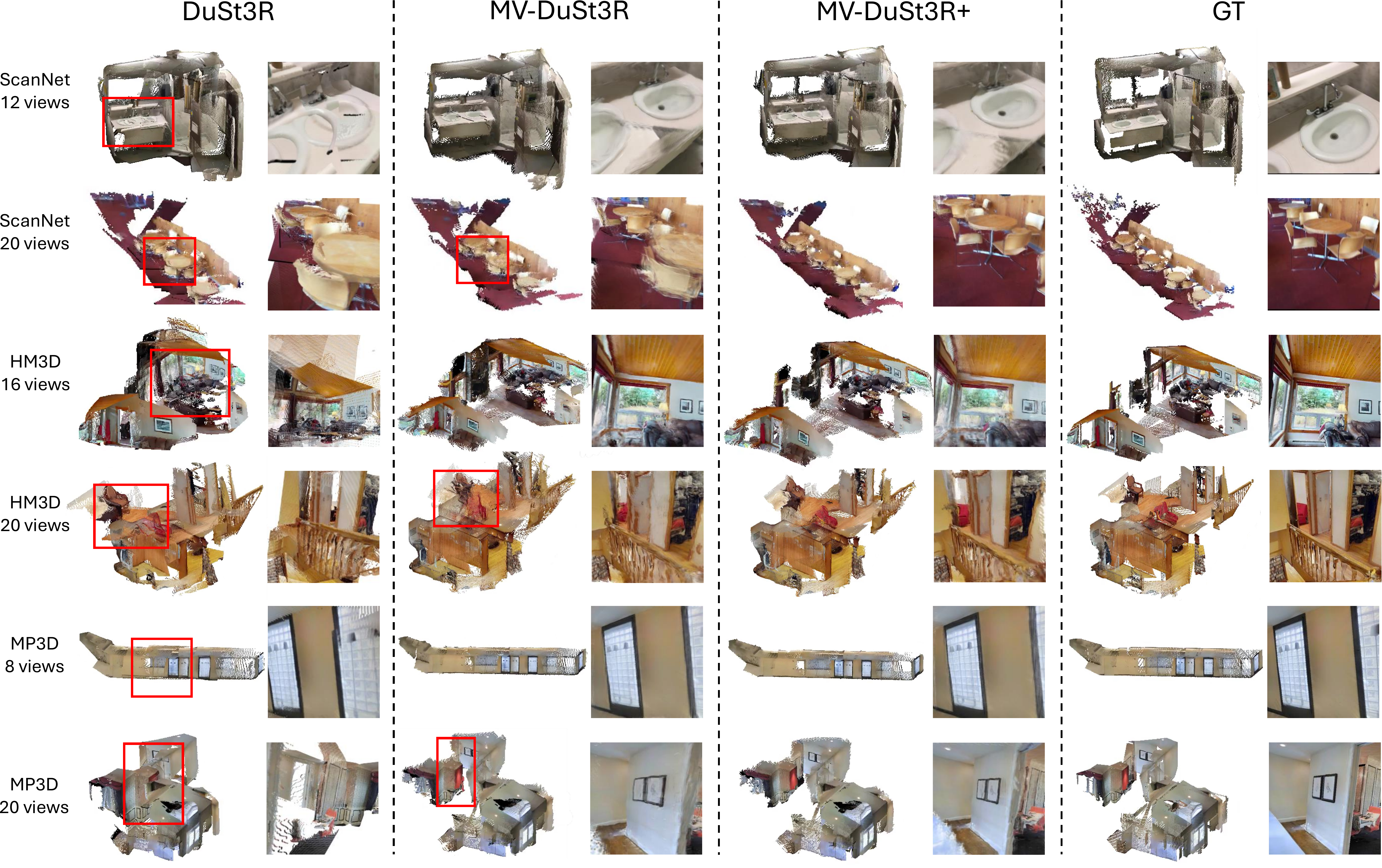}
  \caption{\textbf{MVS reconstruction and NVS qualitative results.} We show one method in each column, which includes the reconstructed pointcloud and 1 rendered new view. Incorrectly reconstructed geometry is highlighted in \textcolor{red}{red} boxes. \dustthreer often introduces incorrect pairwise reconstructions when the scene has multiple objects with similar appearance (\eg, windows, chairs, doors), which can not be recovered by the global optimization. \mvdustthreer is more robust overall but still sometimes fails to reconstruct geometry accurately in regions far away from the reference view, while \mvdustthreerp predicts geometry more evenly across the space. 
  }
  \label{fig:recons_results_comp}
\end{figure*}

\noindent\textbf{Results} are shown in~\cref{tab:MVSReconstruction}. In the supervised setting, we compare \dustthreer and our methods on HM3D and \scannet. On HM3D (multi-room scenes), \textbf{\mvdustthreer} consistently outperforms \dustthreer, as the scene size increases and more input views are sampled (from 4 to 24 views). %
For example, \mvdustthreer reduces \pcddistSHspace by $1.7\times$ and increases \pcdaccSHspace by $1.2\times$ for 4-view input. For 24-view input, \mvdustthreer improves \pcddistSHspace by $2 \times$   and \pcdaccSH~by $5.3\times$. This confirms: as more input views are available, single-stage  \mvdustthreer %
exploits multi-view cues to infer 3D scene geometry better than \dustthreer, which only exploits pairwise stereo cues at a time. Furthermore, \textbf{\mvdustthreerp} substantially improves upon \mvdustthreer, especially when the scene size is large and many more input views are used. With 12-view input, \mvdustthreerp improves \pcddistSHspace by $1.3\times$ and \pcdaccSHspace by $1.2\times$. With 24-view input, the improvements are more significant, with a $1.6\times$ lower \pcddistSHspace and a $1.8\times$ higher \pcdaccSH. The multi-path architecture enables \mvdustthreerp to more effectively  fuse multi-view cues across  different choices of reference frames, which holistically improves the scene geometry reconstruction in all input views. For a qualitative comparisons, see~\cref{fig:recons_results_comp}. For zero-shot evaluation on the challenging MP3D data, all methods have worse results. However, both of our methods consistently improve upon \dustthreer across different numbers of input views.

In all settings, Spann3R performs  worse than all other methods, and often fails to reconstruct a scene from  sparse  views, a setting which is more challenging than the  sequential video frames used for evaluation in the original paper.

\begin{table}[t]
\vspace{-0.3cm}
    \centering
    \small

    \resizebox{0.9\columnwidth}{!}{
    \setlength{\tabcolsep}{2pt}
        \begin{tabular}{c|c|c|ccc|ccc|ccc}
            & \multirow{2}{*}{Method}& \multirow{2}{*}{GO} & \multicolumn{3}{c|}{HM3D} & \multicolumn{3}{c|}{ScanNet} & \multicolumn{3}{c}{MP3D} \\\cline{4-12}
            &  & & RRE $\downarrow$ & RTE $\downarrow$ & mAE $\downarrow$ & RRE $\downarrow$ & RTE $\downarrow$ & mAE $\downarrow$ & RRE $\downarrow$ & RTE $\downarrow$ & mAE $\downarrow$\\
            \specialrule{.2em}{.1em}{.1em}
            \multirow{5}{*}{\rotatebox[origin=c]{90}{4 views}}
            & \dustthreer & \checkmark & 2.4 & 3.1 & 12.5 & 3.0 & 20.0 & 30.7 & 3.5 & 3.8 & 13.3 \\
            & \mvdustthreer & $\times$ & 1.5 & 1.5 & 5.5 & 2.3 & 16.8 & 27.0 & 1.2 & 1.0 & 5.4 \\
            & \mvdustthreerp & $\times$ & \cellcolor[rgb]{0.999, 0.8, 0.8}{1.2} & \cellcolor[rgb]{0.999, 0.8, 0.8}{1.1} & \cellcolor[rgb]{0.999, 0.8, 0.8}{4.9} & \cellcolor[rgb]{0.999, 0.8, 0.8}{1.4} & \cellcolor[rgb]{0.999, 0.8, 0.8}{16.1} & \cellcolor[rgb]{0.999, 0.8, 0.8}{26.2} & \cellcolor[rgb]{0.999, 0.8, 0.8}{0.8} & \cellcolor[rgb]{0.999, 0.8, 0.8}{0.8} & \cellcolor[rgb]{0.999, 0.8, 0.8}{4.6} 
            \\
            \cline{2-12}
            & {\color{lightgray}$\text{\mvdustthreer}_\text{oracle}$} & $\times$ & \lgr{0.0} & \lgr{0.1} & \lgr{2.8} & \lgr{0.9} & \lgr{7.0} & \lgr{18.9} & \lgr{0.1} & \lgr{0.1} & \lgr{2.9} \\
            & {\color{lightgray}$\text{\mvdustthreerp}_\text{oracle}$} & $\times$ & \lgr{0.0} & \lgr{0.0} & \lgr{2.4} & \lgr{0.9} & \lgr{6.8} & \lgr{18.7} & \lgr{0.1} & \lgr{0.0} & \lgr{2.4} 
            \\
            \midrule 
            \multirow{5}{*}{\rotatebox[origin=c]{90}{12 views}} & \dustthreer & \checkmark & 3.7 & 8.3 & 20.1 & 4.6 & 22.6 & 34.2 & 4.5 & 8.4 & 19.8 \\
            & \mvdustthreer & $\times$ & 1.5 & 2.6 & 8.4 & 3.7 & 14.7 & 26.1 & 1.6 & 2.6 & 8.2 \\
            & \mvdustthreerp & $\times$ & \cellcolor[rgb]{0.999, 0.8, 0.8}{0.6} & \cellcolor[rgb]{0.999, 0.8, 0.8}{1.2} & \cellcolor[rgb]{0.999, 0.8, 0.8}{5.2} & \cellcolor[rgb]{0.999, 0.8, 0.8}{2.5} & \cellcolor[rgb]{0.999, 0.8, 0.8}{11.6} & \cellcolor[rgb]{0.999, 0.8, 0.8}{22.9} & \cellcolor[rgb]{0.999, 0.8, 0.8}{0.5} & \cellcolor[rgb]{0.999, 0.8, 0.8}{1.0} & \cellcolor[rgb]{0.999, 0.8, 0.8}{4.9} 
            \\
            \cline{2-12}
            & {\color{lightgray}$\text{\mvdustthreer}_\text{oracle}$} & $\times$ & \lgr{0.4} & \lgr{0.6} & \lgr{4.9} & \lgr{1.7} & \lgr{7.8} & \lgr{20.2} & \lgr{0.6} & \lgr{0.7} & \lgr{5.1} \\
            & {\color{lightgray}$\text{\mvdustthreerp}_\text{oracle}$} & $\times$ & \lgr{0.3} & \lgr{0.3} & \lgr{3.4} & \lgr{1.7} & \lgr{6.0} & \lgr{17.9} & \lgr{0.3} & \lgr{0.3} & \lgr{3.3} 
            \\
            \midrule 
            \multirow{5}{*}{\rotatebox[origin=c]{90}{24 views}} & \dustthreer & \checkmark & 8.8 & 18.1 & 30.9 & 8.1 & 26.6 & 38.9 & 10.0 & 18.2 & 30.5 \\
            & \mvdustthreer & $\times$ & 8.9 & 12.8 & 23.7 & 8.2 & 21.9 & 34.2 & 8.2 & 11.1 & 21.4 \\
            & \mvdustthreerp & $\times$ & \cellcolor[rgb]{0.999, 0.8, 0.8}{3.0} & \cellcolor[rgb]{0.999, 0.8, 0.8}{6.5} & \cellcolor[rgb]{0.999, 0.8, 0.8}{15.8} & \cellcolor[rgb]{0.999, 0.8, 0.8}{4.6} & \cellcolor[rgb]{0.999, 0.8, 0.8}{16.7} & \cellcolor[rgb]{0.999, 0.8, 0.8}{29.4} & \cellcolor[rgb]{0.999, 0.8, 0.8}{3.3} & \cellcolor[rgb]{0.999, 0.8, 0.8}{6.0} & \cellcolor[rgb]{0.999, 0.8, 0.8}{14.6} 
            \\
            \cline{2-12}
            & {\color{lightgray}$\text{\mvdustthreer}_\text{oracle}$} & $\times$ & \lgr{3.2} & \lgr{4.4} & \lgr{14.7} & \lgr{3.4} & \lgr{13.1} & \lgr{26.7} & \lgr{3.4} & \lgr{4.2} & \lgr{14.0} 
            \\
            & {\color{lightgray}$\text{\mvdustthreerp}_\text{oracle}$} & $\times$ & \lgr{1.4} & \lgr{2.4} & \lgr{11.1} & \lgr{2.6} & \lgr{9.9} & \lgr{23.7} & \lgr{1.8} & \lgr{2.4} & \lgr{10.6} 
            \\
            \bottomrule 
            
        \end{tabular}
    }
    \caption{\textbf{Multi-View Pose Estimation results}. Metrics are reported in percent $\%$.}
    \label{tab:MVPE}
\end{table}

\subsection{Multi-View Pose Estimation}
For both baseline and our methods, we estimate the relative camera pose for all pairs of input views from a given set of input views. We use the Weiszfeld algorithm~\citep{weiszfeld} to estimate camera intrinsics, and RANSAC~\citep{fischler1981random} with PnP~\citep{epnp} to estimate camera pose (see appendix for more details).

\noindent\textbf{Baselines.} We compare with other pose-free methods including \dustthreer and PoseDiffusion~\citep{posediffusion}, a recent
diffusion based method for camera pose estimation. 

\noindent\textbf{Metrics.} Prior methods~\citep{Dust3r} report Relative
Rotation Accuracy (\textbf{RRA@15}), Relative Translation Accuracy (\textbf{RTA@15}) under threshold 15 degrees, and mean Average Accuracy (\textbf{mAA@30}) under threshold 30 degrees. 
For clarity, we report Relative Rotation Error (\textbf{RRE@15} = $1.0 - \text{RRA}@15$), Relative Translation Error (\textbf{RTE@15} = $1.0 - \text{RTA}@15$) and mean Average Error (\textbf{mAE@30} = $1.0 - \text{mAA}@30$).

\noindent\textbf{Results.} The comparisons with \dustthreer are presented in~\cref{tab:MVPE}, while comparisons with PoseDiffusion are included in the appendix, as we find PoseDiffusion significantly underperforms other methods on our evaluation datasets. As shown in~\cref{tab:MVPE}, in the supervised setting on HM3D, our method \mvdustthreer achieves a $2.3\times$ lower \maespace for 4-view input, and a $1.3\times$ lower \maespace for 24-view input, compared with \dustthreer. \mvdustthreerp performs best, achieving a $2.6\times$ lower \maespace for 4-view input, and a $2.0\times$ lower \maespace for 24-view input, compared with \dustthreer. 
For another supervised setting ScanNet and the zero-shot MP3D, our method \mvdustthreerp performs best, while \mvdustthreer consistently outperforms \dustthreer under all circumstances.

\subsection{Novel View Synthesis}
\label{sec:resultsNVS}

\noindent\textbf{Metrics.} Following prior works~\citep{splatt3r, pixelsplat, instantsplat}, we report Peak Signal-to-Noise Ratio (\textbf{PSNR}), Structural Similarity Index Measure (\textbf{SSIM})~\citep{ssim}, and Learned Perceptual Image Patch Similarity (\textbf{LPIPS})~\citep{lpips}. 

\noindent\textbf{Baseline.} We compare with a \dustthreer-based baseline, which generates per-pixel Gaussian parameters as follows. We use the pointmap predicted by \dustthreer as the Gaussian center, use pixel RGB color $I^v$ as the  color, a constant $0.001$ for the scale factor $S^{v,m}$, an identity transform, 1.0 for opacity, and  spherical harmonics with zero-degree.
See appendix for more details on rendering. %

\noindent\textbf{Results.} As shown in~\cref{tab:NVS}, \mvdustthreer improves upon the \dustthreer baseline across all evaluation datasets under all choices of input views. The improvements are also confirmed qualitatively in~\cref{fig:recons_results_comp}: the novel views synthesized by \mvdustthreer  better infer 3D geometry of objects and background (\eg, walls, ceiling). \mvdustthreerp further improves in challenging situations, such as a scene with multiple close-by objects of similar appearance (\eg, chairs). As an example, consider the \scannetspace scene with 20 views in \cref{fig:recons_results_comp}. Using multiple  reference views, and fusing features computed in different model paths help resolve ambiguity in inferring the spatial relations between input views.

\begin{table}[t]
    \centering
    \small
    \vspace{-0.3cm}
    \resizebox{0.95\columnwidth}{!}{
    \setlength{\tabcolsep}{2pt}
        \begin{tabular}{c|c|c|ccc|ccc|ccc}
            & \multirow{2}{*}{Method}& \multirow{2}{*}{GO} & \multicolumn{3}{c|}{HM3D} & \multicolumn{3}{c|}{ScanNet} & \multicolumn{3}{c}{MP3D} \\\cline{4-12}
            &  & & PSNR $\uparrow$ & SSIM $\uparrow$& LPIPS $\downarrow$ & PSNR $\uparrow$ & SSIM $\uparrow$& LPIPS $\downarrow$ & PSNR $\uparrow$ & SSIM $\uparrow$& LPIPS $\downarrow$  \\
            \specialrule{.2em}{.1em}{.1em}
            \multirow{5}{*}{\rotatebox[origin=c]{90}{4 views}} & \dustthreer & \checkmark & 16.0 & 5.0 & 3.7 & 17.0 & 6.0 & 3.0 & 15.5 & 4.6 & 4.0\\
            & \mvdustthreer & $\times$ &  19.9 & 6.0 & 2.0 & 21.9 & 7.1 & 1.6 & 19.6 & 5.8 & 2.1\\
            & \mvdustthreerp & $\times$ & \cellcolor[rgb]{0.999, 0.8, 0.8}{20.2} & \cellcolor[rgb]{0.999, 0.8, 0.8}{6.1} & \cellcolor[rgb]{0.999, 0.8, 0.8}{1.9} & \cellcolor[rgb]{0.999, 0.8, 0.8}{22.2} & \cellcolor[rgb]{0.999, 0.8, 0.8}{7.1} & \cellcolor[rgb]{0.999, 0.8, 0.8}{1.5} & \cellcolor[rgb]{0.999, 0.8, 0.8}{19.9} & \cellcolor[rgb]{0.999, 0.8, 0.8}{5.9} & \cellcolor[rgb]{0.999, 0.8, 0.8}{2.0}\\
            \cline{2-12}
            & {\color{lightgray}$\text{\mvdustthreer}_\text{oracle}$} & $\times$ & \lgr{21.0} & \lgr{6.5} & \lgr{1.6} & \lgr{22.8} & \lgr{7.4} & \lgr{1.4} & \lgr{20.6} & \lgr{6.2} & \lgr{1.8}\\
            & {\color{lightgray}$\text{\mvdustthreerp}_\text{oracle}$} & $\times$ & \lgr{21.4} & \lgr{6.6} & \lgr{1.5} & \lgr{23.0} & \lgr{7.4} & \lgr{1.4} & \lgr{21.0} & \lgr{6.3} & \lgr{1.7}\\\midrule 
            \multirow{5}{*}{\rotatebox[origin=c]{90}{12 views}} & \dustthreer & \checkmark & 15.1 & 4.4 & 4.9 & 16.3 & 5.4 & 3.6 & 14.7 & 4.0 & 5.3\\
            & \mvdustthreer & $\times$ & 18.9 & 5.6 & 2.7 & 20.1 & 6.5 & 2.2 & 18.4 & 5.3 & 2.8\\
            & \mvdustthreerp & $\times$ & \cellcolor[rgb]{0.999, 0.8, 0.8}{19.4} & \cellcolor[rgb]{0.999, 0.8, 0.8}{5.8} & \cellcolor[rgb]{0.999, 0.8, 0.8}{2.4} & \cellcolor[rgb]{0.999, 0.8, 0.8}{20.4} & \cellcolor[rgb]{0.999, 0.8, 0.8}{6.6} & \cellcolor[rgb]{0.999, 0.8, 0.8}{2.1} & \cellcolor[rgb]{0.999, 0.8, 0.8}{19.0} & \cellcolor[rgb]{0.999, 0.8, 0.8}{5.5} & \cellcolor[rgb]{0.999, 0.8, 0.8}{2.6}
            \\
            \cline{2-12}
            & {\color{lightgray}$\text{\mvdustthreer}_\text{oracle}$} & $\times$ & \lgr{19.9} & \lgr{5.9} & \lgr{2.2} & \lgr{21.0} & \lgr{6.8} & \lgr{1.9} & \lgr{19.3} & \lgr{5.6} & \lgr{2.4}\\
            & {\color{lightgray}$\text{\mvdustthreerp}_\text{oracle}$} & $\times$ & \lgr{20.4} & \lgr{6.1} & \lgr{2.0} & \lgr{21.3} & \lgr{6.8} & \lgr{1.8} & \lgr{19.9} & \lgr{5.8} & \lgr{2.2}\\\midrule 
            \multirow{5}{*}{\rotatebox[origin=c]{90}{24 views}} & \dustthreer & \checkmark & 14.3 & 4.2 & 5.6 & 15.2 & 5.0 & 4.2 & 13.8 & 3.6 & 6.1\\
            & \mvdustthreer & $\times$ & 17.8 & 5.3 & 3.6 & 18.4 & 6.0 & 2.9 & 17.3 & 4.9 & 3.8\\
            & \mvdustthreerp & $\times$ & \cellcolor[rgb]{0.999, 0.8, 0.8}{18.4} & \cellcolor[rgb]{0.999, 0.8, 0.8}{5.4} & \cellcolor[rgb]{0.999, 0.8, 0.8}{3.2} & \cellcolor[rgb]{0.999, 0.8, 0.8}{18.6} & \cellcolor[rgb]{0.999, 0.8, 0.8}{6.0} & \cellcolor[rgb]{0.999, 0.8, 0.8}{2.8} & \cellcolor[rgb]{0.999, 0.8, 0.8}{17.9} & \cellcolor[rgb]{0.999, 0.8, 0.8}{5.1} & \cellcolor[rgb]{0.999, 0.8, 0.8}{3.5}
            \\
            \cline{2-12}
            & {\color{lightgray}$\text{\mvdustthreer}_\text{oracle}$} & $\times$ & \lgr{18.5} & \lgr{5.5} & \lgr{3.1} & \lgr{19.3} & \lgr{6.2} & \lgr{2.5} & \lgr{18.0} & \lgr{5.1} & \lgr{3.3}\\
            & {\color{lightgray}$\text{\mvdustthreerp}_\text{oracle}$} & $\times$ & \lgr{19.0} & \lgr{5.6} & \lgr{2.9} & \lgr{19.4} & \lgr{6.2} & \lgr{2.5} & \lgr{18.5} & \lgr{5.3} & \lgr{3.1}\\\bottomrule
        \end{tabular}
    }
    \caption{\textbf{Novel View Synthesis results}. For clarity, we scale up metrics SSIM and LPIPS by $10\times$.}
    \label{tab:NVS}
\end{table}

\subsection{Scene Reconstruction Time}
We compare the time of MVS reconstruction in~\cref{tab:MVSReconstruction}. Our single-stage feed-forward networks entirely run on a GPU, without Global Optimization (GO). Compared with \dustthreer, our \mvdustthreer runs $48\times$ to $78\times$ faster than \dustthreer, while the more performant \mvdustthreerp runs $8\times$ to $14\times$ faster, when considering 4 to 24 input views. \textit{\mvdustthreerp reconstructs 24-view input for scenes of average size $17.9$ $m^2$ on HM3D and $37.3$ $m^2$ on MP3D in less than 2 seconds}.

\subsection{Ablation Studies}

\begin{table}[t]
    \centering

    \resizebox{0.95\columnwidth}{!}{
    \setlength{\tabcolsep}{2pt}
        \begin{tabular}{c|c|ccc|ccc|ccc}
            \multirow{2}{*}{Test} & \multirow{2}{*}{Training recipe}  & \multicolumn{3}{c|}{MVS Reconstruction} & \multicolumn{3}{c|}{MVPE} & \multicolumn{3}{c}{NVS} \\\cline{3-11}
              &  & ND $\downarrow$ & DAc $\uparrow$ & CD $\downarrow$ & RRE $\downarrow$ & RTE $\downarrow$ & mAE $\downarrow$ & PSNR $\uparrow$ & SSIM $\uparrow$ & LPIPS $\downarrow$ \\
              \specialrule{.2em}{.1em}{.1em}
            \multirow{3}{*}{\rotatebox[origin=c]{90}{4 views}} & 1-stage, 4 views & 1.0 & 94.4 & 1.7(1.2) & 0.9 & 0.9 & 2.6 & 20.7 & 6.3 & \cellcolor[rgb]{0.999, 0.8, 0.8}1.7 \\

            & 1-stage, 8 views  & 1.0 & 95.2 & 1.5(0.9) & 1.2 & 1.1 & 4.9 & 20.2 & 6.1 & 1.9 \\

            & 2-stage, mixed views &   \cellcolor[rgb]{0.999, 0.8, 0.8}0.9 & \cellcolor[rgb]{0.999, 0.8, 0.8}95.5 & \cellcolor[rgb]{0.999, 0.8, 0.8}1.5(0.8) & \cellcolor[rgb]{0.999, 0.8, 0.8}0.8 & \cellcolor[rgb]{0.999, 0.8, 0.8}0.7 & \cellcolor[rgb]{0.999, 0.8, 0.8}2.0 & \cellcolor[rgb]{0.999, 0.8, 0.8}20.7 & \cellcolor[rgb]{0.999, 0.8, 0.8}6.3 & 1.8 \\
            \midrule
            \multirow{3}{*}{\rotatebox[origin=c]{90}{12 views}} & 1-stage, 4 views & 6.3 & 0.3 & 23.8(18.2) & 15.1 & 17.5 & 34.1 & 16.5 & 4.9 & 4.4 \\

            & 1-stage, 8 views  & 1.2 & 91.5 & 1.8(0.7) & 0.6 & 1.2 & 5.2 & 19.4 & 5.8 & 2.4 \\

            & 2-stage, mixed views & \cellcolor[rgb]{0.999, 0.8, 0.8}1.2 & \cellcolor[rgb]{0.999, 0.8, 0.8}92.2 & \cellcolor[rgb]{0.999, 0.8, 0.8}1.5(1.0) & \cellcolor[rgb]{0.999, 0.8, 0.8}0.4 & \cellcolor[rgb]{0.999, 0.8, 0.8}0.8 & \cellcolor[rgb]{0.999, 0.8, 0.8}3.8 & \cellcolor[rgb]{0.999, 0.8, 0.8}19.5 & \cellcolor[rgb]{0.999, 0.8, 0.8}5.9 & \cellcolor[rgb]{0.999, 0.8, 0.8}2.2 \\
            \midrule
            
            \multirow{3}{*}{\rotatebox[origin=c]{90}{24 views}} & 1-stage, 4 views & 17.7 & 0.0 & 81.4(55.5) & 45.5 & 47.4 & 63.2 & 14.5 & 4.6 & 6.2 \\

            & 1-stage, 8 views  & 2.1 & 64.5 & 3.9(2.0) & 3.0 & 6.5 & 15.8 & 18.4 & 5.4 & 3.2 \\

            & 2-stage, mixed views &   \cellcolor[rgb]{0.999, 0.8, 0.8}1.7 & \cellcolor[rgb]{0.999, 0.8, 0.8}81.4 & \cellcolor[rgb]{0.999, 0.8, 0.8}2.6(1.3) & \cellcolor[rgb]{0.999, 0.8, 0.8}1.4 & \cellcolor[rgb]{0.999, 0.8, 0.8}3.0 & \cellcolor[rgb]{0.999, 0.8, 0.8}9.1 & \cellcolor[rgb]{0.999, 0.8, 0.8}19.1 & \cellcolor[rgb]{0.999, 0.8, 0.8}5.7 & \cellcolor[rgb]{0.999, 0.8, 0.8}2.7 \\
            \bottomrule
        \end{tabular}
    }
    \caption{Impact of  $\#$ of input views at training time on the performance of \mvdustthreerp on the HM3D evaluation set. }
    \label{tab:studyTrainNumberViews}
\end{table}

\begin{table}[t]
\vspace{-0.2cm}
    \centering
    \resizebox{0.85\columnwidth}{!}{
    \setlength{\tabcolsep}{2pt}
        \begin{tabular}{c|c|c|ccc|ccc|ccc}
            & \multirow{2}{*}{Method}& \multirow{2}{*}{GS} &\multicolumn{3}{c|}{HM3D} & \multicolumn{3}{c|}{ScanNet} & \multicolumn{3}{c}{MP3D} \\\cline{4-12}
            &  & & ND $\downarrow$ & DAc $\uparrow$ & CD $\downarrow$ & ND $\downarrow$ & DAc $\uparrow$ & CD $\downarrow$ & ND $\downarrow$ & DAc $\uparrow$ & CD $\downarrow$  \\
            \specialrule{.2em}{.1em}{.1em}
            \multirow{4}{*}{\rotatebox[origin=c]{90}{4 views}} & \multirow{2}{*}{\mvdustthreer}  & $\times$ & 1.1 & \cellcolor[rgb]{0.999, 0.8, 0.8}93.5 & \cellcolor[rgb]{0.999, 0.8, 0.8}1.9(1.4) & 1.0 & 93.3 & \cellcolor[rgb]{0.999, 0.8, 0.8}2.0(0.4) & 2.6 & 61.6 & 25.4(4.9)\\
            &  & \checkmark & \cellcolor[rgb]{0.999, 0.8, 0.8}1.1 & 92.2 & 2.0(1.1) & \cellcolor[rgb]{0.999, 0.8, 0.8}1.0 & \cellcolor[rgb]{0.999, 0.8, 0.8}93.3 & 2.0(0.4) & \cellcolor[rgb]{0.999, 0.8, 0.8}2.5 & \cellcolor[rgb]{0.999, 0.8, 0.8}62.4 & \cellcolor[rgb]{0.999, 0.8, 0.8}25.3(4.1)\\ \cline{2-12}
            & \multirow{2}{*}{\mvdustthreerp} & $\times$ & \cellcolor[rgb]{0.999, 0.8, 0.8}0.9 & \cellcolor[rgb]{0.999, 0.8, 0.8}95.2 & 1.5(1.1) & \cellcolor[rgb]{0.999, 0.8, 0.8}0.8 & \cellcolor[rgb]{0.999, 0.8, 0.8}94.8 & \cellcolor[rgb]{0.999, 0.8, 0.8}1.4(0.4) & \cellcolor[rgb]{0.999, 0.8, 0.8}2.2 & \cellcolor[rgb]{0.999, 0.8, 0.8}68.5 & \cellcolor[rgb]{0.999, 0.8, 0.8}18.9(3.5)\\
            &  & \checkmark & 1.0 & 95.2 & \cellcolor[rgb]{0.999, 0.8, 0.8}1.5(0.9) & 0.8 & 94.9 & 1.5(0.3) & 2.2 & 68.0 & 19.9(3.4)\\\midrule
            \multirow{4}{*}{\rotatebox[origin=c]{90}{24 views}} & \multirow{2}{*}{\mvdustthreer} & $\times$ & \cellcolor[rgb]{0.999, 0.8, 0.8}3.3 & 37.6 & 10.0(5.1) & \cellcolor[rgb]{0.999, 0.8, 0.8}2.2 & \cellcolor[rgb]{0.999, 0.8, 0.8}75.8 & \cellcolor[rgb]{0.999, 0.8, 0.8}2.7(0.7) & 6.4 & 11.0 & 43.3(13.0)\\
            &  & \checkmark  &3.4 & \cellcolor[rgb]{0.999, 0.8, 0.8}36.7 & \cellcolor[rgb]{0.999, 0.8, 0.8}10.0(3.5) & 2.2 & 75.2 & 2.7(0.9) & \cellcolor[rgb]{0.999, 0.8, 0.8}6.3 & \cellcolor[rgb]{0.999, 0.8, 0.8}12.2 & \cellcolor[rgb]{0.999, 0.8, 0.8}38.6(13.9)\\ \cline{2-12}
            & \multirow{2}{*}{\mvdustthreerp} & $\times$ & \cellcolor[rgb]{0.999, 0.8, 0.8}2.1 & \cellcolor[rgb]{0.999, 0.8, 0.8}68.1 & 4.4(2.5) & \cellcolor[rgb]{0.999, 0.8, 0.8}1.4 & \cellcolor[rgb]{0.999, 0.8, 0.8}84.9 & \cellcolor[rgb]{0.999, 0.8, 0.8}1.5(0.5) & 4.3 & 26.5 & 22.6(5.6)\\
            &  & \checkmark & 2.1 & 64.5 & \cellcolor[rgb]{0.999, 0.8, 0.8}3.9(2.0) & 1.6 & 81.2 & 1.7(0.7) & \cellcolor[rgb]{0.999, 0.8, 0.8}4.3 & \cellcolor[rgb]{0.999, 0.8, 0.8}26.7 & \cellcolor[rgb]{0.999, 0.8, 0.8}22.0(5.9)\\\bottomrule 
            
        \end{tabular}
    }
    \vspace{-0.2cm}
    \caption{Impact of adding Gaussian (GS) heads on MVS reconstruction performance on HM3D, \scannet, and MP3D datasets.}
    \label{tab:studyGShead}
\vspace{-0.4cm}
\end{table}

\noindent\textbf{Number of input views at training time.} We compare 1-stage and 2-stage trained \mvdustthreerp on the HM3D evaluation set. %
For 1-stage training, we choose the first 4 or 8 views of the trajectory. For 2-stage training, we finetune upon \mvdustthreerp's 1-stage 8-view training, by using a mixed set of inputs with the number of views uniformly sampled between 4 and 12. As shown in~\cref{tab:studyTrainNumberViews}, 1-stage training on 4 views doesn't generalize well to more views, 1-stage training on 8 views performs decent, and 2-stage training  outperforms 1-stage training on almost all tasks on HM3D. See appendix for more 2-stage results. %

\noindent\textbf{Upper bound performance of our networks}. We study oracle performance  if the best reference view is chosen based on groundtruth. %
For \mvdustthreer, we consider all input views as reference view candidates, and manually select the one with best MVS reconstruction (\textbf{$\text{\mvdustthreer}_\text{oracle}$}). For \mvdustthreerp,  we choose the model path with best MVS reconstruction (\textbf{$\text{\mvdustthreerp}_\text{oracle}$}). As shown in \cref{tab:MVSReconstruction,tab:MVPE,tab:NVS}, 
the performance gap between \mvdustthreerp and $\text{\mvdustthreerp}_\text{oracle}$ is significantly smaller than that between \mvdustthreer and $\text{\mvdustthreer}_\text{oracle}$. This validates our multi-path \mvdustthreerp architecture. %

\noindent\textbf{Impact of adding Gaussian head on MVS reconstruction performance}. In~\cref{tab:studyGShead}, we compare \mvdustthreer and \mvdustthreerp models with and without  Gaussian heads, %
while keeping other settings identical. %
As shown in~\cref{tab:studyGShead}, adding Gaussian heads does not significantly improve or degrade performance of our models on MVS reconstruction.

\section{Conclusion}
We propose fast single-stage networks \mvdustthreer and \mvdustthreerp to reconstruct  scenes from up to 24 input views in one feed-forward pass without requiring camera intrinsics and poses. We extensively evaluate results on 3 datasets in both supervised and zero-shot settings, and confirm  compelling results and efficiency over prior art.

\clearpage
\newpage
\beginappendix

\beginsupplement

\begin{figure*}[t]
  \centering
  \includegraphics[width=1.0\textwidth]{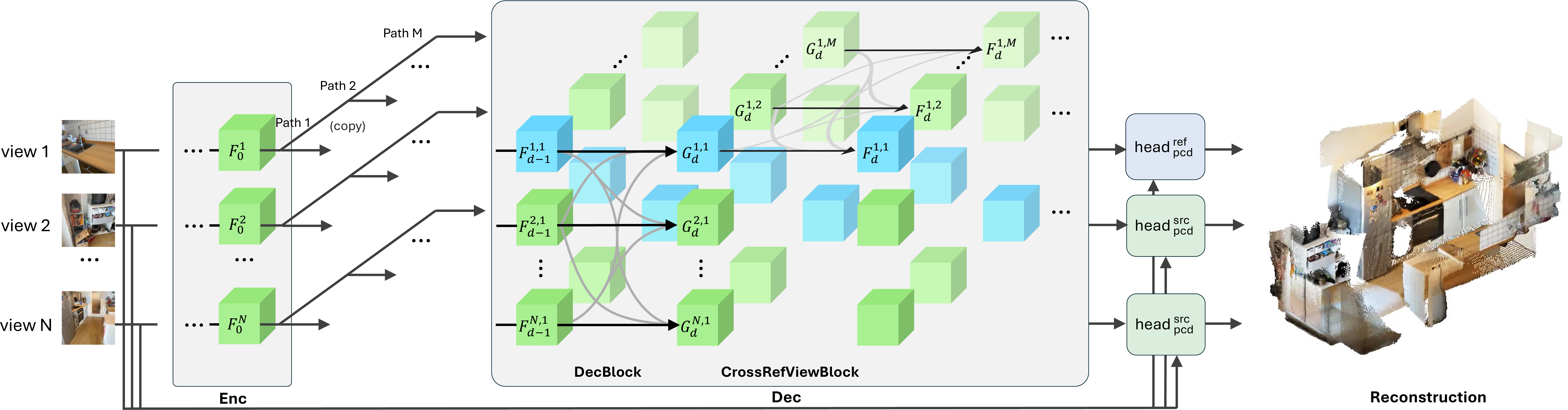}
  \caption{Full pipeline of \mvdustthreerp~ without Gaussian heads. Extra skip connections are sent to the regression head for fine-grained correction. 
  }
  \label{fig:ppl_full}
\end{figure*}

\section{\mvdustthreer(+) for Novel View Synthesis}
\label{sec:nvs_supp}
Below we present more implementation details in order to extend \mvdustthreerp to the NVS task. Note, \mvdustthreer can be viewed as a special case of \mvdustthreerp with the number of reference view candidates $M=1$.

\noindent\textbf{Training Recipe.}
During training, for a chosen reference view $r^m \in R$, we perform differentiable splatting-based rendering~\citep{3dGaussian} to generate rendering predictions for a set of target views $T=\{t^k\}_{k=1}^{N + N'}$, including both $N$ input views and $N'$ novel views. We transform the Gaussian parameters predicted for input view $v$ in the coordinate system of the reference view $r^m$ into  the coordinate system of a target view $t^k$. For this we use the groundtruth camera poses $P^{m}$ of the reference view and $P^k$ of the target view. Formally, the Gaussian centers $X^{v,m}$ are transformed as follows:
\begin{equation}
\hat{X}^{v, m\rightarrow k} = P^{k} (P^m)^{-1} \left(\frac{\bar{z}}{z} X^{v,m}\right).
\end{equation}
Note, we account for scale ambiguity between predicted and groundtruth scene, as described in Section 3.2 of the main paper, by scaling the Gaussian center by $\bar{z}/{z}$. 
We use $\hat{X}^{v, m\rightarrow k}$ to denote the transformed  coordinates. Other Gaussian parameters, including scaling factor $S^{v, m}$ and rotation quaternions $q^{v, m}$, are scaled or transformed accordingly into $\hat{S}^{v, m}$ and $q^{v, m\rightarrow k}$. For each target view $t^k$ and each reference view candidate $r^m$, a splatting-based rendering $\hat{I}^{k, m}$ is generated using all $N$ per-input-view pointmaps predicted from the path $m$ in the multi-path model. Formally, 
\begin{equation}
\hat{I}^{k, m} =\texttt{Rendering}(Q^{1,m\rightarrow k},\dots,Q^{N,m\rightarrow k}),
\end{equation}
where $Q^{v,m\rightarrow k}=\{I^v, \hat{X}^{v, m\rightarrow k}, \hat{S}^{v, m},q^{v, m\rightarrow k}, \alpha^{v, m}\}$ denotes the transformed pixel-aligned Gaussian parameters of input view $v$.

Following prior approaches~\citep{Splatterimage, pixelsplat, Mvsplat}, we use a weighted sum of an $\ell_2$ pixel difference loss and the perceptual similarity loss LPIPS~\citep{lpips} as the rendering loss $\lrender$ to train the Gaussian heads. Formally, 
\begin{equation}
\lrender=\frac{1}{|T||M|}\!\!\!\sum_{(k,m)}\!\!\!\left(\left\| \hat{I}^{k, m} - I^{k} \right\|_2^2 + \gamma\text{LPIPS}(\hat{I}^{k, m}, I^{k})\right),
\end{equation}
where $\gamma$ controls the weight of the LPIPS loss. The final loss to train our models with Gaussian heads includes both the confidence-aware pointmap regression loss and the rendering loss, \ie,
\begin{equation}
\lall=\lconf + \delta\lrender.
\end{equation}
Here, $\delta$ is a weight to balance the two losses.

\noindent\textbf{Model Inference during evaluation.} The inference of \mvdustthreerp with Gaussian heads largely follows the \mvdustthreerp model inference. A subset of $M$ input views are selected as reference views, and the per-view pixel-aligned 3D Gaussian predictions are computed using the heads in the 1st path of the multi-path model. 
To generate a rendering at a novel view during evaluation,  we transform the predicted Gaussian parameters from the 1st reference view to the novel view based on their groundtruth poses, and run splatting-based rendering.

\section{Implementation Details}

\subsection{Network Architecture}

\begin{table}[t]
    \centering
    \small

    \resizebox{0.9\columnwidth}{!}{
    \setlength{\tabcolsep}{2pt}
        \begin{tabular}{c|c|c}
            Module & \dustthreer & \mvdustthreerp \\
            \specialrule{.2em}{.1em}{.1em}
            \multirow{2}{*}{$\texttt{Enc}$} & $\texttt{Conv2d}$(in=3, out=1024, kernel=16, stride=16) & $\texttt{Conv2d}$(in=3, out=1024, kernel=16, stride=16) \\
            & $\texttt{EncBlock}$(embed\_dim=1024, n\_head=16) * 16 & $\texttt{EncBlock}$(embed\_dim=1024, n\_head=16) * 16 \\\midrule
            \multirow{3}{*}{$\texttt{Dec}$} & $\texttt{Linear}$(in=1024, out=768) & $\texttt{Linear}$(in=1024, out=768) \\
            & $\texttt{DecBlock}^\text{ref/src}$(embed\_dim=768, n\_head=12) * 12 & $\texttt{DecBlock}^\text{ref/src}$(embed\_dim=768, n\_head=12) * 12 \\
            & - & \textcolor{red}{$\texttt{CrossRefViewBlock}$(embed\_dim=768, n\_head=12) * 12} \\\midrule
            \multirow{6}{*}{$\texttt{Head}_\text{pcd}$} & $\texttt{Linear}$(in=768, out=768) & $\texttt{Linear}$(in=768, out=768) \\
            & $\texttt{PixelShuffle}$(patch\_size=16) & $\texttt{PixelShuffle}$(patch\_size=16) \\
            & - & \textcolor{red}{$\texttt{Conv2d}$(in=6, out=256, kernel=3, stride=1)} \\
            & - & \textcolor{red}{$\texttt{Conv2d}$(in=256, out=256, kernel=5, stride=1)} \\
            & - & \textcolor{red}{$\texttt{Conv2d}$(in=256, out=256, kernel=5, stride=1)} \\
            & - & \textcolor{red}{$\texttt{Conv2d}$(in=256, out=3, kernel=3, stride=1)} \\\midrule
            \multirow{2}{*}{$\texttt{Head}_\text{3DGS}$} & - & \textcolor{red}{$\texttt{Linear}$(in=768, out=768)} \\
            & - & \textcolor{red}{$\texttt{PixelShuffle}$(patch\_size=16)} \\
            \bottomrule 
            
        \end{tabular}
    }
    \vspace{-0.2cm}
    \caption{\textbf{Architecture Comparisons}. New parameters in \mvdustthreerp are highlighted in \textcolor{red}{red} color.}
    \label{tab:supp_param}
\end{table}

An overview of the \mvdustthreerp model is provided in \cref{fig:ppl_full}. Inspired by \dustthreer, our new $\texttt{CrossRefViewBlock}$ shares the same block architecture as the $\texttt{DecBlock}$, which includes a cross-attention block, a self-attention block and a MLP. Its parameters are randomly initialized except that zero-initialization is used for the last layer in the cross attention block, the self attention block and the MLP. Our Gaussian prediction heads share the same structure as \dst's pointmap prediction head. However, our pointmap prediction head differs and has slightly more trainable parameters %
to improve  accuracy. 

For clarity, we slightly abuse the notation and remove the superscript of the input view and the reference view below to present the head $\texttt{Head}_\text{pcd}$. We start with the linear head design in \dustthreer, which consists of a \texttt{Linear} layer followed by a \texttt{PixelShuffle} layer to restore the original resolution. Differently, we improve the vanilla head by adding a skip connection from the input view to the head output to restore more high-resolution details in the final pointmap prediction. The skip connection is implemented as a small $\texttt{ConvNet}$, which consists  of multiple stride-1 convolutional layers:
\begin{align}
X^{v}_\text{c}&=\texttt{PixelShuffle}(\texttt{Linear}(F^{v}_D)),\\
X^{v}&=\texttt{ConvNet}(X^{v}_\text{c}, I^v) + X^{v}_\text{c}.
\end{align}
Here, $X^{v}_\text{c}$ is the coarse point map and the \texttt{ConvNet} includes 4 sequential stride-1 convolutional layers with kernal size 3, 5, 5, and 3 respectively. See \cref{tab:supp_param} for more details and a comparison of network architectures and hyperparameters.

\subsection{Model Training Recipe}
We implement our approach in PyTorch~\citep{pytorch}. We train our models for $100$ epochs using $150K$ trajectories per epoch,  a learning rate of $1.5\mathrm{e}{-4}$ with a cosine schedule, and Adam optimizer~\citep{adam}. Training takes 180 hours. The hyperparameters in the loss functions are $\beta=0.2$, $\gamma=1$, and $\delta=1$. The input image resolution is $224 \times 224$. We use the open-sourced weights of \dustthreer\footnote{\url{https://github.com/naver/dust3r?tab=readme-ov-file\#checkpoints}}, trained at the input resolution $224\times224$, to initialize all parameters existing within our models. We do not finetune the \dustthreer model trained at the higher $512$ resolution for two reasons: 1) the lower resolution model is representative enough given its performance on Multi-View depth evaluation is similar to that of the higher resolution model as reported in the original \dustthreer paper; and 2) finetuning at the higher $512$ resolution is computationally costly. We leave scaling of \mvdustthreerp training at the higher $512$ resolution to  future work.

We use the first $N=8$ views of each 10-view training trajectory as input views. Among those $N=8$ input views, we select \textit{one} random view as the reference view when training \mvdustthreer and $M=4$ random reference views when training \mvdustthreerp. The remaining $N'=2$ views are considered as novel views.

\section{Experiments}

\subsection{Datasets}
\label{sec:suppDataset}

\begin{figure*}[t]
  \centering
  \includegraphics[width=1.0\textwidth]{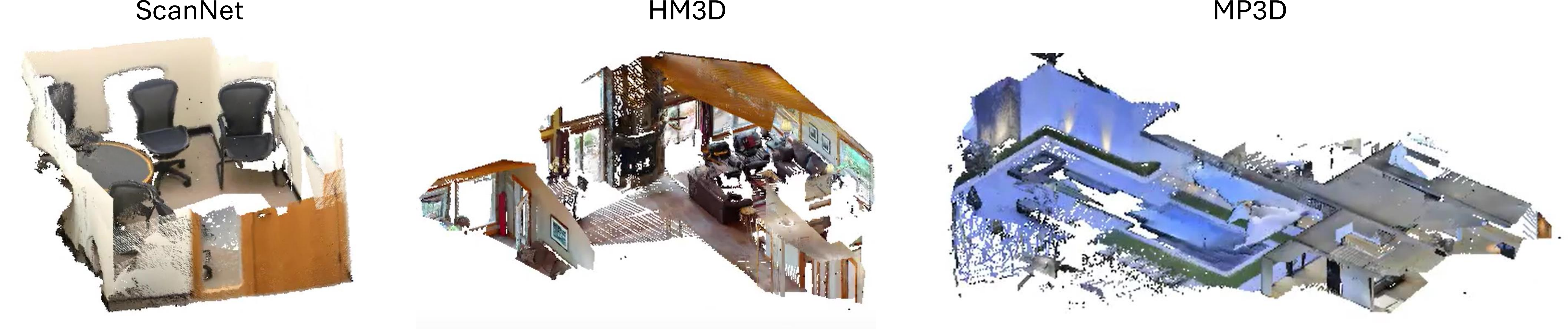}
  \caption{\textbf{Scenes with 20 sampled input views from the evaluation datasets}. \scannetspace scenes are small and only contains a single room. HM3D scenes often cover a larger area and multiple rooms. MP3D scenes are on average the largest with the highest diversity, which often not only cover some indoor region, but also the outdoor pool and garden.  
  }
  \label{fig:scene}
\end{figure*}

In our experiments, we use 
ScanNet~\citep{scannet} (1221 scenes for training, 292 scenes for test), ScanNet++~\citep{scannet++} (360 scenes for training), HM3D~\citep{hm3d} (800 scenes for training, 100 scenes for test), Gibson~\citep{gibson} (500 scenes for training) and MP3D~\citep{chang2017matterport3d} (no scenes for training, 18 scenes for test). Note that our 3 evaluation datasets differ  in  scene size and type. ScanNet scenes are often small-sized single-rooms with low scene diversity. Scenes in MP3D and HM3D are often large-sized multi-room settings with high diversity. MP3D also contains outdoor scenes. 
See~\cref{fig:scene} for a comparison of the evaluation datasets.

\noindent\textbf{Training Trajectories.} 
To sample the training set trajectories, for ScanNet and ScanNet++,  we employ two sets of thresholds: $\text{(}t_{\text{min}}, t_{\text{max}} \text{)} \in \{\text{(}30\%, 70\%\text{)}, \text{(}30\%, 100\%\text{)} \} $. %
For each set of thresholds, we sample $1k$ trajectories of $10$ views per scene, for a total of $3.16M$ trajectories from ScanNet and ScanNet++. %
For HM3D and Gibson, we only deploy the more challenging threshold set $\text{(}t_{\text{min}}, t_{\text{max}} \text{)} = \text{(}30\%, 70\%\text{)}$ because we empirically observe overfitting in the experiments using the other threshold choice. %
Here, we sample 6K trajectories per scene with 10 views each, for a total of 7.8M trajectories. Note that we sample more trajectories because HM3D and Gibson scenes~\citep{hm3d} cover a much larger area than those in ScanNet and ScanNet++.

To generate a trajectory, we first choose as a starting view  either one random view from the predefined trajectories in the  \scannet~ and \scannet++ dataset, or render from a random location in Habitat-Sim~\citep{habitat} for the HM3D, Gibson, and MP3D datasets. Then we iteratively sample a new random view and add it to the trajectory if it meets a criterion (described later). We repeat this step until a total of 10 views are sampled. We now describe  whether a new candidate view $I^i$ meets the criterion. Let's refer to its pointmap as $X^i$ and let $\{X^j\}_{j<i}$ denote the pointmaps of views that are already sampled. %
Using those, we calculate the overlap ratio $O(X^i,X^j)$ of each pair $(X^i,X^j)$. Formally,
\begin{subequations}
\begin{equation}
O(X^i,X^j)=\frac{1}{2}(\texttt{Cov}(X^i, X^j) + \texttt{Cov}(X^j, X^i)),
\end{equation}
\begin{equation}
\texttt{Cov}(X^i, X^j) = \frac{1}{|A|}\sum_{p\in X^i}[\texttt{NearestDis}(p, X^j) < t_\text{c}],
\end{equation}
\end{subequations}
where $|A|$ denotes the number of points in $X_i$ and $t_c=0.0015$ is the threshold on the 3D point distance. 
A new view $X^i$ is selected if the largest overlap ratio $\max_j \{ O(X^i,X^j) \}_j$ is between $t_{\text{min}}$ and $t_{\text{max}}$.

\noindent\textbf{Test Trajectories.} 
For testing, we generate $1K$ trajectories for each dataset %
following the procedure described above. The only difference: we sample $30$-view trajectories, including 24 input views and 6 novel views for NVS evaluation. Those 6 novel views are selected to make sure that the 1st novel view is covered well by input views 1$\sim$4, the 2nd novel view is covered well by input views 5$\sim$8  and so on. To evaluate NVS with  4-view input, we use the extra 1st view for evaluation; for 8-view input, we use the extra 1st and 2nd view for evaluation and so on.

\subsection{Multi-View Stereo Reconstruction}
\label{sec:supp_MVS}

\noindent\textbf{More results of  up to 24 input views}. In \cref{tab:supp_MVS}, we present additional results for input views $N \in \{8, 16, 20\}$. Those results are not presented in the main paper's Table 2. Consistent with the findings in Table 2, Spann3R struggles in all settings on all 3 evaluation datasets. \mvdustthreerp consistently outperforms all other approaches in all settings, while \mvdustthreer substantially outperforms \dustthreer by a  margin.

\begin{table}[t]
    \centering
    \resizebox{0.9\columnwidth}{!}{
    \setlength{\tabcolsep}{2pt}
        \begin{tabular}{c|c|c|ccc|ccc|ccc|c}
            & \multirow{2}{*}{Method}& \multirow{2}{*}{GO} & \multicolumn{3}{c|}{HM3D} & \multicolumn{3}{c|}{ScanNet} & \multicolumn{3}{c}{MP3D} & Time\\\cline{4-12}
            &  & & ND $\downarrow$ & DAc $\uparrow$ & CD $\downarrow$ & ND $\downarrow$ & DAc $\uparrow$ & CD $\downarrow$ & ND $\downarrow$ & DAc $\uparrow$ & CD $\downarrow$ & (sec) \\
            \specialrule{.2em}{.1em}{.1em}
            \multirow{5}{*}{\rotatebox[origin=c]{90}{4 views}}
            & Spann3R & $\times$ & 37.1 & 0.0 & 225(184) & 8.9 & 19.5 & 54.7(50.1) & 42.7 & 0.0 & 248(202) & 0.36\\
            & \dustthreer & \checkmark & 1.9 & 75.1 & 5.6(2.3) & 1.3 & 89.8 & 4.0(0.4) & 3.9 & 41.7 & 40.0(5.3) & 2.42\\
            & \mvdustthreer & $\times$ & 1.1 & 92.2 & 2.0(1.1) & 1.0 & 93.3 & 2.0(0.4) & 2.5 & 62.4 & 25.3(4.1)& \cellcolor[rgb]{0.999, 0.8, 0.8}0.05\\
            & \mvdustthreerp & $\times$ & \cellcolor[rgb]{0.999, 0.8, 0.8} 1.0 & \cellcolor[rgb]{0.999, 0.8, 0.8}95.2 & \cellcolor[rgb]{0.999, 0.8, 0.8}1.5(0.9) & \cellcolor[rgb]{0.999, 0.8, 0.8}0.8 & \cellcolor[rgb]{0.999, 0.8, 0.8}94.9 & \cellcolor[rgb]{0.999, 0.8, 0.8}1.5(0.3) & \cellcolor[rgb]{0.999, 0.8, 0.8}2.2 & \cellcolor[rgb]{0.999, 0.8, 0.8}68.0 & \cellcolor[rgb]{0.999, 0.8, 0.8}19.9(3.4) & 0.29\\\cline{2-13}
            & {\color{lightgray}$\text{\mvdustthreer}_\text{oracle}$} & $\times$ & \lgr{1.0} & \lgr{94.6} & \lgr{1.5(0.7)} & \lgr{0.8} & \lgr{95.5} & \lgr{1.3(0.3)} & \lgr{2.3} & \lgr{66.6} & \lgr{20.7(4.0)}& -\\
            & {\color{lightgray}$\text{\mvdustthreerp}_\text{oracle}$} & $\times$ & \lgr{0.9} & \lgr{96.5} & \lgr{1.4(0.7)} & \lgr{0.7} & \lgr{95.8} & \lgr{1.2(0.2)} & \lgr{2.1} & \lgr{70.6} & \lgr{17.9(3.3)}& -\\\midrule 
            \multirow{5}{*}{\rotatebox[origin=c]{90}{8 views}} 
            & Spann3R & $\times$ & 37.1 & 0.0 & 225(184) & 8.9 & 19.5 & 54.7(50.1) & 42.7 & 0.0 & 248(202) & 0.79\\
            & \dustthreer & \checkmark & 3.0 & 48.5 & 11.9(2.7) & 1.7 & 86.4 & 4.2(0.6) & 5.1 & 21.5 & 40.3(5.7) & 4.49\\
            & \mvdustthreer & $\times$ & 1.4 & 87.4 & 2.5(1.2) & 1.2 & 90.4 & 2.0(0.4) & 2.9 & 53.8 & 21.1(5.1) & \cellcolor[rgb]{0.999, 0.8, 0.8}0.10\\
            & \mvdustthreerp & $\times$ & \cellcolor[rgb]{0.999, 0.8, 0.8}1.1 & \cellcolor[rgb]{0.999, 0.8, 0.8}93.5 & \cellcolor[rgb]{0.999, 0.8, 0.8}2.9(0.8) & \cellcolor[rgb]{0.999, 0.8, 0.8}1.0 & \cellcolor[rgb]{0.999, 0.8, 0.8}92.1 & \cellcolor[rgb]{0.999, 0.8, 0.8}1.5(0.4) & \cellcolor[rgb]{0.999, 0.8, 0.8}2.3 & \cellcolor[rgb]{0.999, 0.8, 0.8}62.0 & \cellcolor[rgb]{0.999, 0.8, 0.8}15.1(4.2) & 0.56\\\cline{2-13}
            & {\color{lightgray}$\text{\mvdustthreer}_\text{oracle}$} & $\times$ & \lgr{1.1} & \lgr{92.4} & \lgr{1.7(1.1)} & \lgr{0.9} & \lgr{92.8} & \lgr{1.3(0.4)} & \lgr{2.5} & \lgr{59.4} & \lgr{16.7(4.0)}& -\\
            & {\color{lightgray}$\text{\mvdustthreerp}_\text{oracle}$} & $\times$ & \lgr{1.0} & \lgr{95.3} & \lgr{1.8(0.8)} & \lgr{0.9} & \lgr{93.2} & \lgr{1.3(0.4)} & \lgr{2.2} & \lgr{65.4} & \lgr{14.0(4.2)}& -\\\midrule 
            \multirow{5}{*}{\rotatebox[origin=c]{90}{12 views}}
            & Spann3R & $\times$ & 32.6 &  0.0 & 125(113) & 9.1 & 16.3 & 36.6(31.2) & 35.0 & 0.0 & 138(112) & 1.34\\
            & \dustthreer & \checkmark &3.9 & 30.7 & 18.1(3.4) & 1.9 & 82.6 & 4.1(0.6) & 6.6 & 12.0 & 49.6(8.3) & 8.28\\
            & \mvdustthreer & $\times$ & 1.6 & 79.5 & 3.0(1.2) & 1.4 & 86.8 & 2.3(0.8) & 3.4 & 41.3 & 22.6(5.5) & \cellcolor[rgb]{0.999, 0.8, 0.8}0.15\\
            & \mvdustthreerp & $\times$ & \cellcolor[rgb]{0.999, 0.8, 0.8}{1.2} & \cellcolor[rgb]{0.999, 0.8, 0.8}{91.5} & \cellcolor[rgb]{0.999, 0.8, 0.8}{1.8(0.7)} &\cellcolor[rgb]{0.999, 0.8, 0.8}{ 1.2} & \cellcolor[rgb]{0.999, 0.8, 0.8}{88.4} & \cellcolor[rgb]{0.999, 0.8, 0.8}{1.8(0.7)} & \cellcolor[rgb]{0.999, 0.8, 0.8}{2.6} & \cellcolor[rgb]{0.999, 0.8, 0.8}{55.0} & \cellcolor[rgb]{0.999, 0.8, 0.8}{15.1(3.8)} & 0.89\\\cline{2-13}
            & {\color{lightgray}$\text{\mvdustthreer}_\text{oracle}$} & $\times$ & \lgr{1.3} & \lgr{88.8} & \lgr{1.8(0.9)} & \lgr{1.0} & \lgr{90.6} & \lgr{1.3(0.7)} & \lgr{2.9} & \lgr{51.3} & \lgr{16.4(4.0)} & -\\
            & {\color{lightgray}$\text{\mvdustthreerp}_\text{oracle}$} & $\times$ & \lgr{1.1} & \lgr{94.8} & \lgr{1.4(0.7)} & \lgr{1.0} & \lgr{90.9} & \lgr{1.3(0.5)} & \lgr{2.5} & \lgr{59.8} & \lgr{13.6(3.5)}& -\\\midrule 
            \multirow{5}{*}{\rotatebox[origin=c]{90}{16 views}}
            & Spann3R & $\times$ & 37.1 & 0.0 & 225(184) & 8.9 & 19.5 & 54.7(50.1) & 42.7 & 0.0 & 248(202) & 1.73\\
            & \dustthreer & \checkmark & 4.8 & 19.8 & 21.4(4.1) & 2.1 & 78.1 & 3.9(0.9) & 8.1 & 7.1 & 60.5(11.1) & 13.07\\
            & \mvdustthreer & $\times$ & 2.0 & 67.5 & 4.2(2.1) & 1.7 & 84.3 & 2.3(1.3) & 4.2 & 29.2 & 26.1(8.5) & \cellcolor[rgb]{0.999, 0.8, 0.8}0.21 \\
            & \mvdustthreerp & $\times$ & \cellcolor[rgb]{0.999, 0.8, 0.8}1.5 & \cellcolor[rgb]{0.999, 0.8, 0.8}85.5 & \cellcolor[rgb]{0.999, 0.8, 0.8}2.3(1.5) & \cellcolor[rgb]{0.999, 0.8, 0.8}1.4 & \cellcolor[rgb]{0.999, 0.8, 0.8}86.6 & \cellcolor[rgb]{0.999, 0.8, 0.8}1.8(0.8) & \cellcolor[rgb]{0.999, 0.8, 0.8}3.2 & \cellcolor[rgb]{0.999, 0.8, 0.8}44.1 & \cellcolor[rgb]{0.999, 0.8, 0.8}17.3(4.4)& 1.19\\\cline{2-13}
            & {\color{lightgray}$\text{\mvdustthreer}_\text{oracle}$} & $\times$ & \lgr{1.5} & \lgr{82.5} & \lgr{2.1(1.5)} & \lgr{1.2} & \lgr{89.3} & \lgr{1.3(0.7)} & \lgr{3.3} & \lgr{39.6} & \lgr{17.4(4.5)}& -\\
            & {\color{lightgray}$\text{\mvdustthreerp}_\text{oracle}$} & $\times$ & \lgr{1.3} & \lgr{90.2} & \lgr{1.7(1.1)} & \lgr{1.2} & \lgr{89.6} & \lgr{1.3(0.7)} & \lgr{2.8} & \lgr{51.4} & \lgr{14.2(3.2)}& -\\\midrule 
            \multirow{5}{*}{\rotatebox[origin=c]{90}{20 views}}
            & Spann3R & $\times$ & 37.1 & 0.0 & 225(184) & 8.9 & 19.5 & 54.7(50.1) & 42.7 & 0.0 & 248(202) & 2.19\\
            & \dustthreer & \checkmark & 5.7 & 12.0 & 27.1(4.9) & 2.3 & 74.3 & 4.6(0.9) & 9.7 & 4.1 & 69.2(11.1) & 19.59\\
            & \mvdustthreer & $\times$ & 2.7 & 51.6 & 6.7(2.5) & 1.9 & 79.3 & 2.5(1.0) & 5.2 & 19.8 & 33.1(10.5) & \cellcolor[rgb]{0.999, 0.8, 0.8}0.28 \\
            & \mvdustthreerp & $\times$ & \cellcolor[rgb]{0.999, 0.8, 0.8}1.8 & \cellcolor[rgb]{0.999, 0.8, 0.8}76.9 & \cellcolor[rgb]{0.999, 0.8, 0.8}3.1(1.2) & \cellcolor[rgb]{0.999, 0.8, 0.8}1.5 & \cellcolor[rgb]{0.999, 0.8, 0.8}84.0 & \cellcolor[rgb]{0.999, 0.8, 0.8}1.8(0.7) & \cellcolor[rgb]{0.999, 0.8, 0.8}3.6 & \cellcolor[rgb]{0.999, 0.8, 0.8}34.2 & \cellcolor[rgb]{0.999, 0.8, 0.8}18.1(6.7)& 1.54\\\cline{2-13}
            & {\color{lightgray}$\text{\mvdustthreer}_\text{oracle}$} & $\times$ & \lgr{1.8} & \lgr{72.5} & \lgr{2.7(1.3)} & \lgr{1.3} & \lgr{86.4} & \lgr{1.4(0.6)} & \lgr{3.7} & \lgr{30.8} & \lgr{17.3(5.8)}& -\\
            & {\color{lightgray}$\text{\mvdustthreerp}_\text{oracle}$} & $\times$ & \lgr{1.5} & \lgr{84.3} & \lgr{2.1(1.0)} & \lgr{1.3} & \lgr{86.8} & \lgr{1.3(0.5)} & \lgr{3.2} & \lgr{41.2} & \lgr{13.0(4.8)}& -\\\midrule 
            \multirow{5}{*}{\rotatebox[origin=c]{90}{24 views}}
            & Spann3R & $\times$ & 41.7 & 0.0 & 139(121) & 11.4 & 1.6 & 37.4(35.5) & 46.6 & 0.0 & 151(121) & 2.73\\
            & \dustthreer & \checkmark & 6.8 & 7.3 & 32.4(5.2) & 2.4 & 72.6 & 5.1(1.0) & 11.4 & 2.5 & 80.9(14.3) & 27.21\\
            & \mvdustthreer & $\times$ & 3.4 & 36.7 & 10.0(3.5) & 2.2 & 75.2 & 2.7(0.9) & 6.3 & 12.2 & 38.6(13.9) & \cellcolor[rgb]{0.999, 0.8, 0.8}0.35\\
            & \mvdustthreerp & $\times$ & \cellcolor[rgb]{0.999, 0.8, 0.8}{2.1} & \cellcolor[rgb]{0.999, 0.8, 0.8}{64.5} & \cellcolor[rgb]{0.999, 0.8, 0.8}{3.9(2.0)} & \cellcolor[rgb]{0.999, 0.8, 0.8}{1.6 }& \cellcolor[rgb]{0.999, 0.8, 0.8}{81.2} & \cellcolor[rgb]{0.999, 0.8, 0.8}{1.7(0.7)} & \cellcolor[rgb]{0.999, 0.8, 0.8}{4.3} & \cellcolor[rgb]{0.999, 0.8, 0.8}{26.7} & \cellcolor[rgb]{0.999, 0.8, 0.8}{22.0(5.9)} & 1.97\\\cline{2-13}
            & {\color{lightgray}$\text{\mvdustthreer}_\text{oracle}$} & $\times$ & \lgr{2.1} & \lgr{58.9} & \lgr{3.5(2.1)} & \lgr{1.4} & \lgr{82.9} & \lgr{1.4(0.7)} & \lgr{4.4} & \lgr{22.0} & \lgr{19.9(5.1)}& -\\
            & {\color{lightgray}$\text{\mvdustthreerp}_\text{oracle}$} & $\times$ & \lgr{1.8} & \lgr{77.9} & \lgr{2.6(1.3)} & \lgr{1.3} & \lgr{85.1} & \lgr{1.3(0.6)} & \lgr{3.6} & \lgr{33.1} & \lgr{15.1(4.4)}& -\\\bottomrule 
            
        \end{tabular}
    }
    \caption{\textbf{Additional MVS reconstruction results}. We use the same notation and scales of metrics as  Table 2 in the main paper (the same below). %
    }
    \label{tab:supp_MVS}
\end{table}

\begin{figure*}[t]
  \centering
  \includegraphics[width=1.0\textwidth]{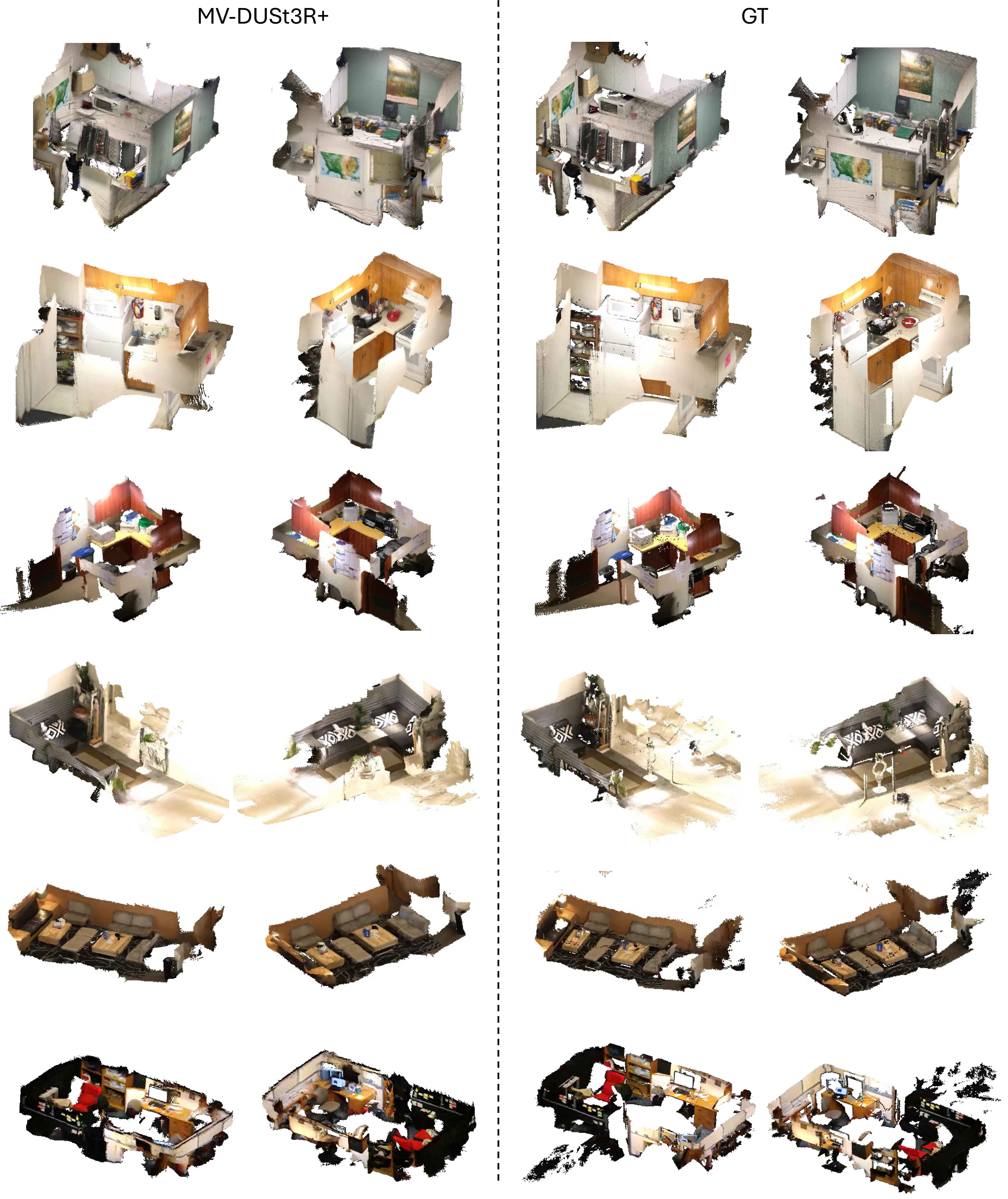}
  \caption{\textbf{MVS reconstruction results with 100-view inputs on \scannet}. $3\%$ of the predicted points with low confidence score are filtered out (the same below). Qualitatively we find the reconstructed scene geometry to be very similar to the groundtruth. This shows that our \mvdustthreerp model, trained with 8-view samples, can generalize remarkably well to 100-view inputs for single-room scenes. The inference time for 100 views is 19.1 seconds.
  }
  \label{fig:supp_100_sc}
\end{figure*}

\begin{figure*}[t]
  \centering
  \includegraphics[width=1.0\textwidth]{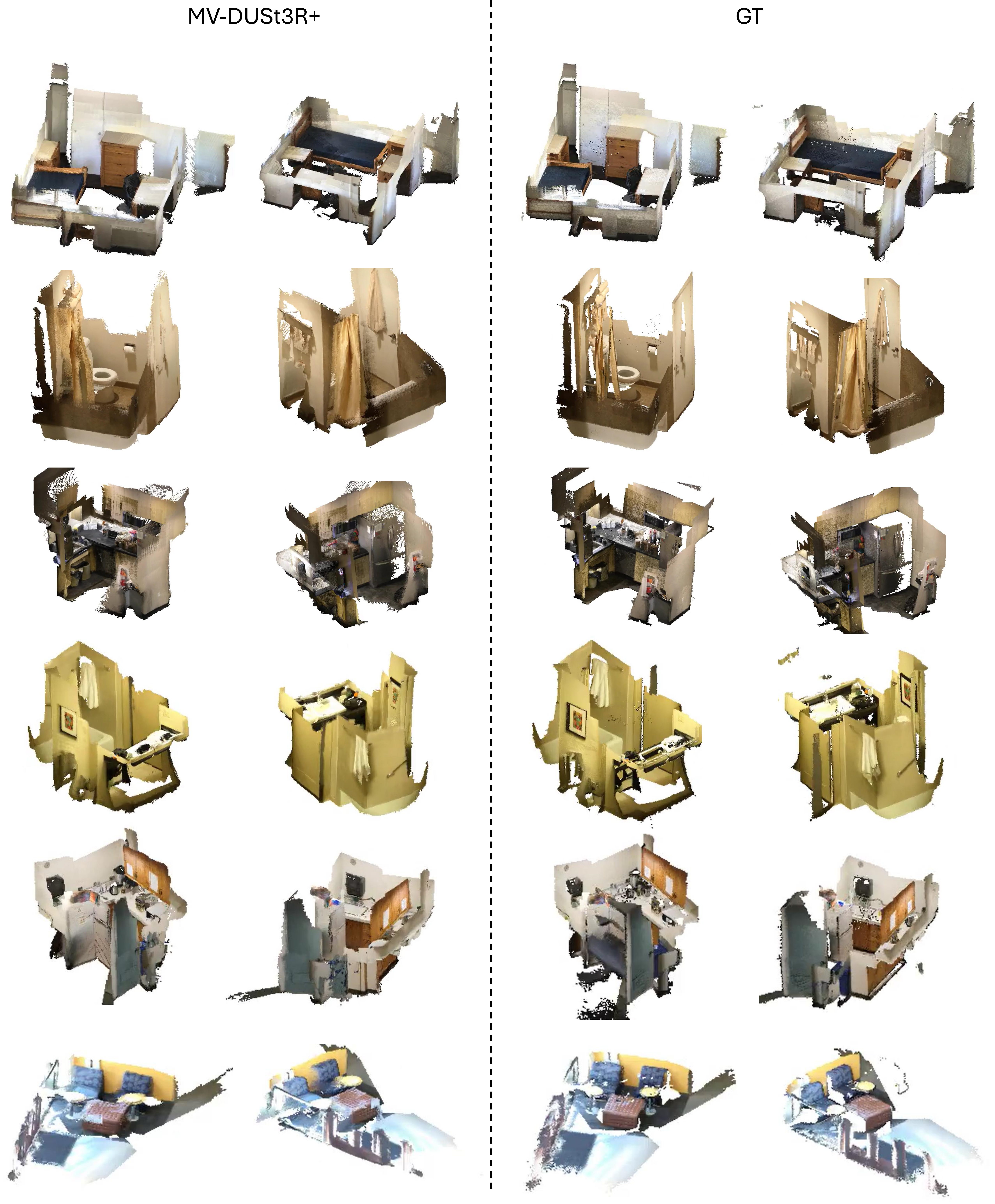}
  \caption{\textbf{Extra MVS reconstruction results with 100-view inputs on \scannet}.
  }
  \label{fig:supp_100_sc2}
\end{figure*}

\begin{figure*}[t]
  \centering
  \includegraphics[width=1.0\textwidth]{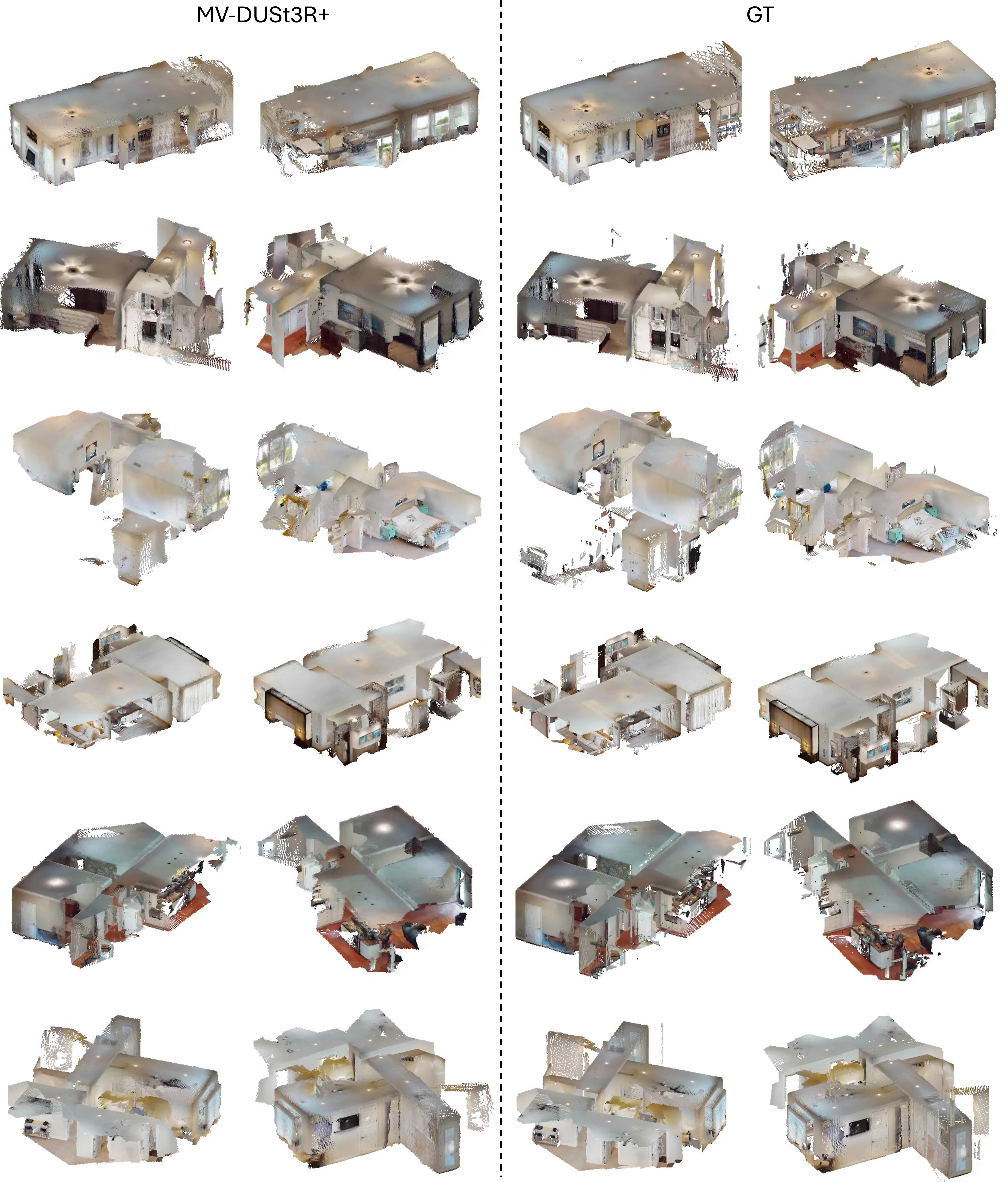}
  \caption{\textbf{MVS reconstruction results with 100-view inputs on HM3D}. \mvdustthreerp still performs well on  more challenging multi-room scenes.
  }
  \label{fig:supp_100_hm3d}
\end{figure*}

\begin{figure*}[t]
  \centering
  \includegraphics[width=1.0\textwidth]{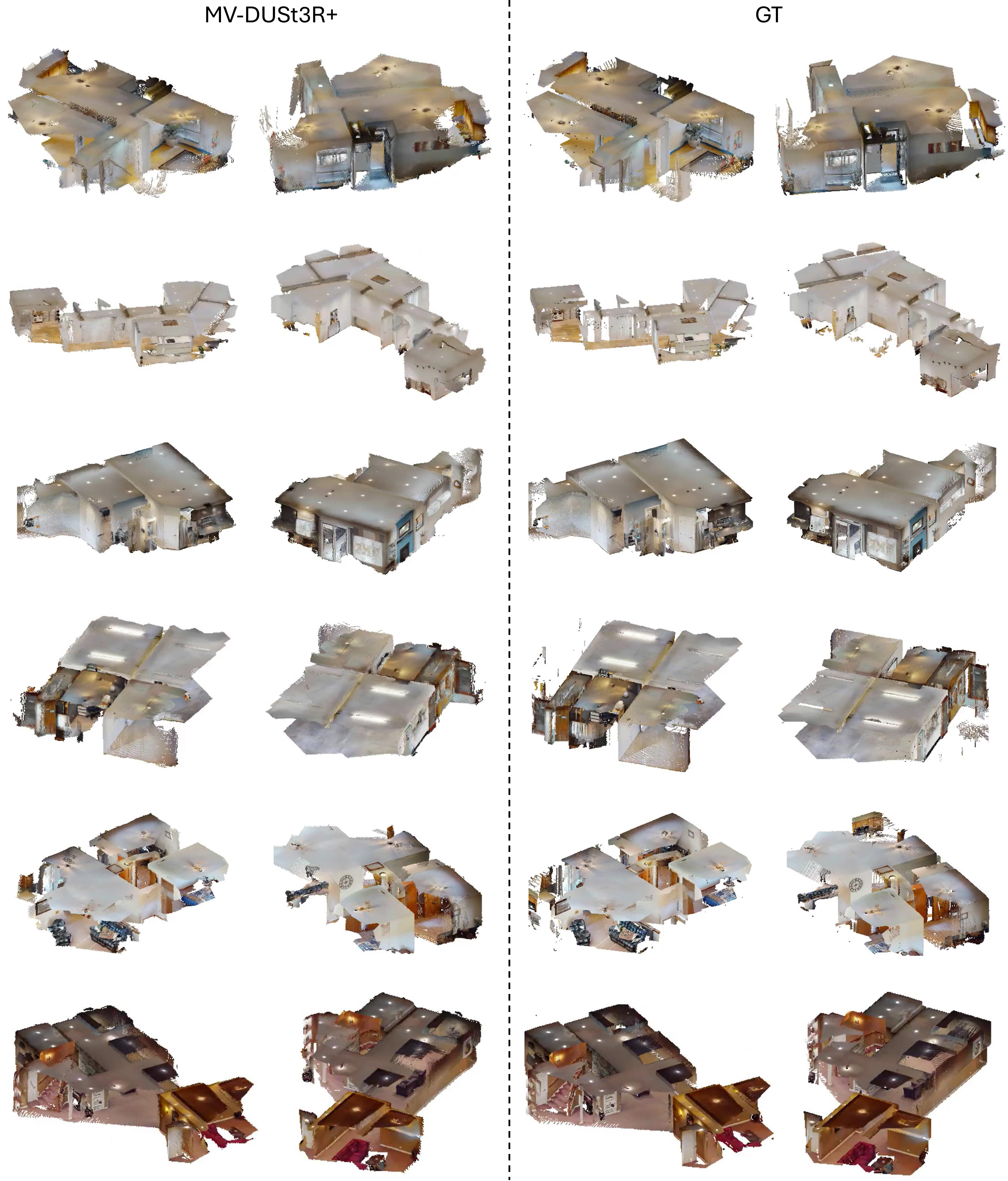}
  \caption{\textbf{Additional MVS reconstruction results with 100-view inputs on HM3D}.}
  \label{fig:supp_100_hm3d2}
\end{figure*}

\noindent\textbf{Results of 100-view input}. Next we validate the generalization performance of \mvdustthreerp using 100 input views while the number of input views at training time is fixed to 8. Specifically,  we present qualitative MVS reconstruction results on \scannetspace (see ~\cref{fig:supp_100_sc} and \cref{fig:supp_100_sc2}) and HM3D (see~\cref{fig:supp_100_hm3d} and \cref{fig:supp_100_hm3d2}). For \scannetspace data, we uniformly sample 100 views from the pre-defined trajectory in the original dataset. For HM3D data, we adopt the  trajectory generation approach described in Section 4.1 of the main paper, but use less restrictive  view overlapping thresholds  $\text{(}t_{\text{min}}, t_{\text{max}} \text{)} = \text{(}50\%, 100\%\text{)}$ to sample a much larger number of input views. 

As shown in~\cref{fig:supp_100_sc} and ~\cref{fig:supp_100_sc2}, \mvdustthreerp generalizes to 100-view input remarkably well on \scannet, and is able to reconstruct a scene geometry close to the groundtruth. On HM3D data, which contains much larger multi-room scenes, our  \mvdustthreerp still performs remarkably well in reconstructing the scene (e.g., layout, walls, ceiling, objects), as shown in~\cref{fig:supp_100_hm3d} and~\cref{fig:supp_100_hm3d2}.

\subsection{Multi-View Pose Estimation}
\label{sec:supp_MVPE}

\begin{table}[t]
    \centering
    \small

    \resizebox{0.8\columnwidth}{!}{
    \setlength{\tabcolsep}{2pt}
        \begin{tabular}{c|c|c|ccc|ccc|ccc}
            & \multirow{2}{*}{Method}& \multirow{2}{*}{GO} & \multicolumn{3}{c|}{HM3D} & \multicolumn{3}{c|}{ScanNet} & \multicolumn{3}{c}{MP3D} \\\cline{4-12}
            &  & & RRE $\downarrow$ & RTE $\downarrow$ & mAE $\downarrow$ & RRE $\downarrow$ & RTE $\downarrow$ & mAE $\downarrow$ & RRE $\downarrow$ & RTE $\downarrow$ & mAE $\downarrow$\\
            \specialrule{.2em}{.1em}{.1em}
            \multirow{5}{*}{\rotatebox[origin=c]{90}{4 views}} & \dustthreer & \checkmark & 2.4 & 3.1 & 12.5 & 3.0 & 20.0 & 30.7 & 3.5 & 3.8 & 13.3 \\
            & \mvdustthreer & $\times$ & 1.5 & 1.5 & 5.5 & 2.3 & 16.8 & 27.0 & 1.2 & 1.0 & 5.4\\
            & \mvdustthreerp & $\times$ & \cellcolor[rgb]{0.999, 0.8, 0.8}1.2 & \cellcolor[rgb]{0.999, 0.8, 0.8}1.1 & \cellcolor[rgb]{0.999, 0.8, 0.8}4.9 & \cellcolor[rgb]{0.999, 0.8, 0.8}1.4 & \cellcolor[rgb]{0.999, 0.8, 0.8}16.1 & \cellcolor[rgb]{0.999, 0.8, 0.8}26.2 & \cellcolor[rgb]{0.999, 0.8, 0.8}0.8 & \cellcolor[rgb]{0.999, 0.8, 0.8}0.8 & \cellcolor[rgb]{0.999, 0.8, 0.8}4.6\\\cline{2-12}
            &  {\color{lightgray}$\text{\mvdustthreer}_\text{oracle}$} & $\times$  & \lgr{0.0} & \lgr{0.1} & \lgr{2.8} & \lgr{0.9} & \lgr{7.0} & \lgr{18.9} & \lgr{0.1} & \lgr{0.1} & \lgr{2.9}\\
            &  {\color{lightgray}$\text{\mvdustthreerp}_\text{oracle}$} & $\times$  & \lgr{0.0} & \lgr{0.0} & \lgr{2.4} & \lgr{0.9} & \lgr{6.8} & \lgr{18.7} & \lgr{0.1} & \lgr{0.0} & \lgr{2.4}\\\midrule 
            \multirow{5}{*}{\rotatebox[origin=c]{90}{8 views}} & \dustthreer & \checkmark & 3.0 & 5.8 & 16.5 & 4.0 & 22.0 & 33.0 & 3.0 & 5.3 & 16.0\\
            & \mvdustthreer & $\times$ & 1.1 & 1.5 & 5.1 & 2.9 & 12.9 & 24.0 & 0.9 & 1.2 & 4.9\\
            & \mvdustthreerp & $\times$ & \cellcolor[rgb]{0.999, 0.8, 0.8}0.6 & \cellcolor[rgb]{0.999, 0.8, 0.8}0.8 & \cellcolor[rgb]{0.999, 0.8, 0.8}2.9 & \cellcolor[rgb]{0.999, 0.8, 0.8}2.0 & \cellcolor[rgb]{0.999, 0.8, 0.8}10.1 & \cellcolor[rgb]{0.999, 0.8, 0.8}21.0 & \cellcolor[rgb]{0.999, 0.8, 0.8}0.4 & \cellcolor[rgb]{0.999, 0.8, 0.8}0.5 & \cellcolor[rgb]{0.999, 0.8, 0.8}2.6\\\cline{2-12}
            &  {\color{lightgray}$\text{\mvdustthreer}_\text{oracle}$} & $\times$  & \lgr{0.1} & \lgr{0.2} & \lgr{2.8} & \lgr{1.6} & \lgr{6.5} & \lgr{18.7} & \lgr{0.1} & \lgr{0.2} & \lgr{2.8}\\
            &  {\color{lightgray}$\text{\mvdustthreerp}_\text{oracle}$} & $\times$  & \lgr{0.1} & \lgr{0.1} & \lgr{1.6} & \lgr{1.4} & \lgr{5.3} & \lgr{16.6} & \lgr{0.1} & \lgr{0.1} & \lgr{1.5}\\\midrule 
            \multirow{5}{*}{\rotatebox[origin=c]{90}{12 views}} & \dustthreer & \checkmark & 3.7 & 8.3 & 20.1 & 4.6 & 22.6 & 34.2 & 4.5 & 8.4 & 19.8\\
            & \mvdustthreer & $\times$ & 1.5 & 2.6 & 8.4 & 3.7 & 14.7 & 26.1 & 1.6 & 2.6 & 8.2\\
            & \mvdustthreerp & $\times$ & \cellcolor[rgb]{0.999, 0.8, 0.8}0.6 & \cellcolor[rgb]{0.999, 0.8, 0.8}1.2 & \cellcolor[rgb]{0.999, 0.8, 0.8}5.2 & \cellcolor[rgb]{0.999, 0.8, 0.8}2.5 & \cellcolor[rgb]{0.999, 0.8, 0.8}11.6 & \cellcolor[rgb]{0.999, 0.8, 0.8}22.9 & \cellcolor[rgb]{0.999, 0.8, 0.8}0.5 & \cellcolor[rgb]{0.999, 0.8, 0.8}1.0 & \cellcolor[rgb]{0.999, 0.8, 0.8}4.9\\\cline{2-12}
            &  {\color{lightgray}$\text{\mvdustthreer}_\text{oracle}$} & $\times$  & \lgr{0.4} & \lgr{0.6} & \lgr{4.9} & \lgr{1.7} & \lgr{7.8} & \lgr{20.2} & \lgr{0.6} & \lgr{0.7} & \lgr{5.1}\\
            &  {\color{lightgray}$\text{\mvdustthreerp}_\text{oracle}$} & $\times$  & \lgr{0.3} & \lgr{0.3} & \lgr{3.4} & \lgr{1.7} & \lgr{6.0} & \lgr{17.9} & \lgr{0.3} & \lgr{0.3} & \lgr{3.3}\\\midrule 
            \multirow{5}{*}{\rotatebox[origin=c]{90}{16 views}} & \dustthreer & \checkmark & 5.0 & 11.3 & 23.6 & 6.3 & 24.6 & 36.3 & 6.1 & 11.5 & 23.6\\
            & \mvdustthreer & $\times$ & 3.0 & 5.2 & 13.0 & 4.9 & 17.2 & 28.9 & 3.1 & 5.0 & 12.5\\
            & \mvdustthreerp & $\times$ & \cellcolor[rgb]{0.999, 0.8, 0.8}1.0 & \cellcolor[rgb]{0.999, 0.8, 0.8}2.5 & \cellcolor[rgb]{0.999, 0.8, 0.8}8.5 & \cellcolor[rgb]{0.999, 0.8, 0.8}3.1 & \cellcolor[rgb]{0.999, 0.8, 0.8}13.4 & \cellcolor[rgb]{0.999, 0.8, 0.8}25.3 & \cellcolor[rgb]{0.999, 0.8, 0.8}1.4 & \cellcolor[rgb]{0.999, 0.8, 0.8}2.4 & \cellcolor[rgb]{0.999, 0.8, 0.8}8.1\\\cline{2-12}
            &  {\color{lightgray}$\text{\mvdustthreer}_\text{oracle}$} & $\times$  & \lgr{0.7} & \lgr{1.1} & \lgr{7.5} & \lgr{2.4} & \lgr{9.8} & \lgr{22.7} & \lgr{1.1} & \lgr{1.3} & \lgr{7.8}\\
            &  {\color{lightgray}$\text{\mvdustthreerp}_\text{oracle}$} & $\times$  & \lgr{0.5} & \lgr{0.6} & \lgr{5.7} & \lgr{2.0} & \lgr{7.6} & \lgr{20.3} & \lgr{0.8} & \lgr{0.8} & \lgr{5.7}\\\midrule 
            \multirow{5}{*}{\rotatebox[origin=c]{90}{20 views}} & \dustthreer & \checkmark & 6.5 & 14.6 & 27.2 & 7.7 & 26.0 & 38.1 & 7.9 & 14.8 & 27.1\\
            & \mvdustthreer & $\times$ & 5.5 & 8.7 & 18.1 & 6.6 & 19.7 & 31.7 & 5.1 & 7.8 & 16.7\\
            & \mvdustthreerp & $\times$ & \cellcolor[rgb]{0.999, 0.8, 0.8}1.6 & \cellcolor[rgb]{0.999, 0.8, 0.8}4.4 & \cellcolor[rgb]{0.999, 0.8, 0.8}12.2 & \cellcolor[rgb]{0.999, 0.8, 0.8}3.8 & \cellcolor[rgb]{0.999, 0.8, 0.8}15.2 & \cellcolor[rgb]{0.999, 0.8, 0.8}27.6 & \cellcolor[rgb]{0.999, 0.8, 0.8}1.9 & \cellcolor[rgb]{0.999, 0.8, 0.8}3.8 & \cellcolor[rgb]{0.999, 0.8, 0.8}11.1\\\cline{2-12}
            &  {\color{lightgray}$\text{\mvdustthreer}_\text{oracle}$} & $\times$  & \lgr{1.7} & \lgr{2.5} & \lgr{10.8} & \lgr{2.7} & \lgr{11.4} & \lgr{24.8} & \lgr{1.8} & \lgr{2.5} & \lgr{10.5}\\
            &  {\color{lightgray}$\text{\mvdustthreerp}_\text{oracle}$} & $\times$  & \lgr{0.7} & \lgr{1.3} & \lgr{8.3} & \lgr{2.3} & \lgr{8.7} & \lgr{22.2} & \lgr{1.0} & \lgr{1.4} & \lgr{7.9}\\\midrule 
            \multirow{5}{*}{\rotatebox[origin=c]{90}{24 views}} & \dustthreer & \checkmark & 8.8 & 18.1 & 30.9 & 8.1 & 26.6 & 38.9 & 10.0 & 18.2 & 30.5\\
            & \mvdustthreer & $\times$ & 8.9 & 12.8 & 23.7 & 8.2 & 21.9 & 34.2 & 8.2 & 11.1 & 21.4\\
            & \mvdustthreerp & $\times$ & \cellcolor[rgb]{0.999, 0.8, 0.8}3.0 & \cellcolor[rgb]{0.999, 0.8, 0.8}6.5 & \cellcolor[rgb]{0.999, 0.8, 0.8}15.8 & \cellcolor[rgb]{0.999, 0.8, 0.8}4.6 & \cellcolor[rgb]{0.999, 0.8, 0.8}16.7 & \cellcolor[rgb]{0.999, 0.8, 0.8}29.4 & \cellcolor[rgb]{0.999, 0.8, 0.8}3.3 & \cellcolor[rgb]{0.999, 0.8, 0.8}6.0 & \cellcolor[rgb]{0.999, 0.8, 0.8}14.6\\\cline{2-12}
            &  {\color{lightgray}$\text{\mvdustthreer}_\text{oracle}$} & $\times$  & \lgr{3.2} & \lgr{4.4} & \lgr{14.7} & \lgr{3.4} & \lgr{13.1} & \lgr{26.7} & \lgr{3.4} & \lgr{4.2} & \lgr{14.0}\\
            &  {\color{lightgray}$\text{\mvdustthreerp}_\text{oracle}$} & $\times$  & \lgr{1.4} & \lgr{2.4} & \lgr{11.1} & \lgr{2.6} & \lgr{9.9} & \lgr{23.7} & \lgr{1.8} & \lgr{2.4} & \lgr{10.6}\\

            \bottomrule 
            
        \end{tabular}
    }
    \caption{\textbf{Additional Multi-View Pose Estimation results}.}
    \label{tab:supp_MVPE}
\end{table}

\noindent\textbf{Camera Intrinsics Estimation}. We assume the camera intrinsics are the same for all the views in the trajectory, and thus only estimate it for the 1st view. At inference time, both \mvdustthreer and \mvdustthreerp predict pointmap $\bar{X}^{1, 1}$ for the 1st reference view $I^1$ in its own camera coordinate system. We further assumes the principal point is in the center of the image plane, and a pixel is square. The remaining unknown camera intrinsic parameter $f_1^{*}$ can be estimated by solving the following minimization problem:
\begin{equation}
 f^*_1 = \arg \min_{f_1} \sum_{(i,j)}C^{1,1}_{i,j}\left\| (i', j') - f_1 \frac{(\bar{X}^{1,1}_{i,j,0}, \bar{X}^{1,1}_{i,j,1})}{\bar{X}^{1,1}_{i,j,2}} \right\|_2^2.    
\end{equation}
Here, $(i', j') = (i - \frac{H}{2}, j - \frac{W}{2})$ denotes the normalized 2D image coordinate.
Following \dustthreer, we use the fast  Weiszfeld algorithm~\citep{weiszfeld} to estimate $f^*_1$.

\noindent\textbf{Camera Pose Estimation}. To estimate the relative camera pose between 2 input views $I^i$ and $I^j$, we first estimate per-view camera pose $P_k^*$ for $k\in\{1,\dots,N\}$, where $P^k \in \mathbb{R}^{4\times4}$. For this we use RANSAC~\citep{fischler1981random} with PnP~\citep{epnp} based on the correspondences between 2D pixels and the corresponding 3D pointmap $\bar{X}^{k, 1}$, predicted by \mvdustthreer or \mvdustthreerp. In practice, for the 1st view $I^1$, the estimated $P^1$ is often slightly different from the ideal identity matrix. After that, we estimate the relative camera pose as $P^j(P^i)^{-1}$.

\noindent\textbf{Additional Results}. In~\cref{tab:supp_MVPE}, we report MVPE results for different numbers of input views. Consistent with the findings in  Table 3 of the main paper, \mvdustthreerp performs  best in all settings on all 3 evaluation datasets, and \mvdustthreer consistently outperforms \dustthreer.

We also report results of PoseDiffusion~\citep{posediffusion} on HM3D and \scannet. We use their Github open-source implementation\footnote{\url{https://github.com/facebookresearch/PoseDiffusion}} and load the checkpoint of the model trained on RealEstate10K data~\citep{RealEstate10K}. As shown in~\cref{tab:supp_posediffusion}, PoseDiffusion struggles on HM3D and \scannet. We hypothesize that this is because PoseDiffusion  is trained on input views with only slightly different camera poses and less diverse indoor scenes (see visualization in the original paper~\citep{posediffusion}). It thus is not able to generalize to our more challenging  HM3D and \scannet~data, where input views are more sparse and the scenes are larger and more diverse.

\begin{table}[t]
    \centering
    \small

    \resizebox{0.5\columnwidth}{!}{
    \setlength{\tabcolsep}{2pt}
        \begin{tabular}{c|ccc|ccc}
            \# of & \multicolumn{3}{c|}{HM3D} & \multicolumn{3}{c}{ScanNet} \\\cline{2-7}
             views  & RRE $\downarrow$ & RTE $\downarrow$ & mAE $\downarrow$ & RRE $\downarrow$ & RTE $\downarrow$ & mAE $\downarrow$ \\
              
            \specialrule{.2em}{.1em}{.1em}
            \rotatebox[origin=c]{0}{4} & 93.3 & 90.0 & 100.0 & 96.7 & 86.7 & 98.9 \\\midrule
            \rotatebox[origin=c]{0}{12} & 94.5 & 94.8 & 98.8 & 97.0 & 98.5 & 99.9 \\\midrule
            \rotatebox[origin=c]{0}{24} & 95.1 & 96.7 & 99.5 & 98.5 & 96.3 & 99.8 \\
            \bottomrule 
            
        \end{tabular}
    }
    \caption{\textbf{PoseDiffusion Evaluation Results}. Metrics are reported in percent $\%$.}
    \label{tab:supp_posediffusion}
\end{table}

\subsection{Novel View Synthesis}

\begin{table}[t]
    \centering
    \small
    \resizebox{0.8\columnwidth}{!}{
    \setlength{\tabcolsep}{2pt}
        \begin{tabular}{c|c|c|ccc|ccc|ccc}
            & \multirow{2}{*}{Method}& \multirow{2}{*}{GO} & \multicolumn{3}{c|}{HM3D} & \multicolumn{3}{c|}{ScanNet} & \multicolumn{3}{c}{MP3D} \\\cline{4-12}
            &  & & PSNR $\uparrow$ & SSIM $\uparrow$& LPIPS $\downarrow$ & PSNR $\uparrow$ & SSIM $\uparrow$& LPIPS $\downarrow$ & PSNR $\uparrow$ & SSIM $\uparrow$& LPIPS $\downarrow$  \\
            \specialrule{.2em}{.1em}{.1em}
            
            \multirow{5}{*}{\rotatebox[origin=c]{90}{4 views}} & \dustthreer & \checkmark & 16.0 & 5.0 & 3.7 & 17.0 & 6.0 & 3.0 & 15.5 & 4.6 & 4.0\\
            & \mvdustthreer & $\times$ & 19.9 & 6.0 & 2.0 & 21.9 & 7.1 & 1.6 & 19.6 & 5.8 & 2.1\\
            & \mvdustthreerp & $\times$ & \cellcolor[rgb]{0.999, 0.8, 0.8}20.2 & \cellcolor[rgb]{0.999, 0.8, 0.8}6.1 & \cellcolor[rgb]{0.999, 0.8, 0.8}1.9 & \cellcolor[rgb]{0.999, 0.8, 0.8}22.2 & \cellcolor[rgb]{0.999, 0.8, 0.8}7.1 & \cellcolor[rgb]{0.999, 0.8, 0.8}1.5 & \cellcolor[rgb]{0.999, 0.8, 0.8}19.9 & \cellcolor[rgb]{0.999, 0.8, 0.8}5.9 & \cellcolor[rgb]{0.999, 0.8, 0.8}2.0\\\cline{2-12}
            &  {\color{lightgray}$\text{\mvdustthreer}_\text{oracle}$} & $\times$  & \lgr{21.0} & \lgr{6.5} & \lgr{1.6} & \lgr{22.8} & \lgr{7.4} & \lgr{1.4} & \lgr{20.6} & \lgr{6.2} & \lgr{1.8}\\
            &  {\color{lightgray}$\text{\mvdustthreerp}_\text{oracle}$} & $\times$  & \lgr{21.4} & \lgr{6.6} & \lgr{1.5} & \lgr{23.0} & \lgr{7.4} & \lgr{1.4} & \lgr{21.0} & \lgr{6.3} & \lgr{1.7}\\\midrule 
            \multirow{5}{*}{\rotatebox[origin=c]{90}{8 views}} & \dustthreer & \checkmark & 15.5 & 4.6 & 4.4 & 16.7 & 5.6 & 3.4 & 15.0 & 4.2 & 4.8\\
            & \mvdustthreer & $\times$ & 19.3 & 5.8 & 2.3 & 21.0 & 6.8 & 1.9 & 18.9 & 5.5 & 2.5\\
            & \mvdustthreerp & $\times$ & \cellcolor[rgb]{0.999, 0.8, 0.8}19.8 & \cellcolor[rgb]{0.999, 0.8, 0.8}6.0 & \cellcolor[rgb]{0.999, 0.8, 0.8}2.1 & \cellcolor[rgb]{0.999, 0.8, 0.8}21.3 & \cellcolor[rgb]{0.999, 0.8, 0.8}6.9 & \cellcolor[rgb]{0.999, 0.8, 0.8}1.8 & \cellcolor[rgb]{0.999, 0.8, 0.8}19.5 & \cellcolor[rgb]{0.999, 0.8, 0.8}5.7 & \cellcolor[rgb]{0.999, 0.8, 0.8}2.2\\\cline{2-12}
            &  {\color{lightgray}$\text{\mvdustthreer}_\text{oracle}$} & $\times$  & \lgr{20.6} & \lgr{6.2} & \lgr{1.9} & \lgr{21.9} & \lgr{7.1} & \lgr{1.6} & \lgr{20.0} & \lgr{6.0} & \lgr{2.0}\\
            &  {\color{lightgray}$\text{\mvdustthreerp}_\text{oracle}$} & $\times$  & \lgr{21.2} & \lgr{6.5} & \lgr{1.7} & \lgr{22.2} & \lgr{7.2} & \lgr{1.6} & \lgr{20.7} & \lgr{6.2} & \lgr{1.8}\\\midrule 
            \multirow{5}{*}{\rotatebox[origin=c]{90}{12 views}} & \dustthreer & \checkmark & 15.1 & 4.4 & 4.9 & 16.3 & 5.4 & 3.6 & 14.7 & 4.0 & 5.3\\
            & \mvdustthreer & $\times$ & 18.9 & 5.6 & 2.7 & 20.1 & 6.5 & 2.2 & 18.4 & 5.3 & 2.8\\
            & \mvdustthreerp & $\times$ & \cellcolor[rgb]{0.999, 0.8, 0.8}19.4 & \cellcolor[rgb]{0.999, 0.8, 0.8}5.8 & \cellcolor[rgb]{0.999, 0.8, 0.8}2.4 & \cellcolor[rgb]{0.999, 0.8, 0.8}20.4 & \cellcolor[rgb]{0.999, 0.8, 0.8}6.6 & \cellcolor[rgb]{0.999, 0.8, 0.8}2.1 & \cellcolor[rgb]{0.999, 0.8, 0.8}19.0 & \cellcolor[rgb]{0.999, 0.8, 0.8}5.5 & \cellcolor[rgb]{0.999, 0.8, 0.8}2.6\\\cline{2-12}
            &  {\color{lightgray}$\text{\mvdustthreer}_\text{oracle}$} & $\times$  & \lgr{19.9} & \lgr{5.9} & \lgr{2.2} & \lgr{21.0} & \lgr{6.8} & \lgr{1.9} & \lgr{19.3} & \lgr{5.6} & \lgr{2.4}\\
            &  {\color{lightgray}$\text{\mvdustthreerp}_\text{oracle}$} & $\times$  & \lgr{20.4} & \lgr{6.1} & \lgr{2.0} & \lgr{21.3} & \lgr{6.8} & \lgr{1.8} & \lgr{19.9} & \lgr{5.8} & \lgr{2.2}\\\midrule 
            \multirow{5}{*}{\rotatebox[origin=c]{90}{16 views}} & \dustthreer & \checkmark & 14.8 & 4.3 & 5.2 & 15.8 & 5.2 & 3.9 & 14.3 & 3.8 & 5.6\\
            & \mvdustthreer & $\times$ & 18.5 & 5.5 & 3.0 & 19.4 & 6.3 & 2.4 & 18.0 & 5.2 & 3.2\\
            & \mvdustthreerp & $\times$ & \cellcolor[rgb]{0.999, 0.8, 0.8}19.0 & \cellcolor[rgb]{0.999, 0.8, 0.8}5.6 & \cellcolor[rgb]{0.999, 0.8, 0.8}2.7 & \cellcolor[rgb]{0.999, 0.8, 0.8}19.6 & \cellcolor[rgb]{0.999, 0.8, 0.8}6.3 & \cellcolor[rgb]{0.999, 0.8, 0.8}2.3 & \cellcolor[rgb]{0.999, 0.8, 0.8}18.6 & \cellcolor[rgb]{0.999, 0.8, 0.8}5.3 & \cellcolor[rgb]{0.999, 0.8, 0.8}2.9\\\cline{2-12}
            &  {\color{lightgray}$\text{\mvdustthreer}_\text{oracle}$} & $\times$  & \lgr{19.3} & \lgr{5.8} & \lgr{2.5} & \lgr{20.4} & \lgr{6.6} & \lgr{2.1} & \lgr{18.8} & \lgr{5.4} & \lgr{2.7}\\
            &  {\color{lightgray}$\text{\mvdustthreerp}_\text{oracle}$} & $\times$  & \lgr{19.8} & \lgr{5.9} & \lgr{2.3} & \lgr{20.5} & \lgr{6.6} & \lgr{2.1} & \lgr{19.4} & \lgr{5.6} & \lgr{2.5}\\\midrule 
            \multirow{5}{*}{\rotatebox[origin=c]{90}{20 views}} & \dustthreer & \checkmark & 14.6 & 4.2 & 5.4 & 15.5 & 5.1 & 4.1 & 14.1 & 3.7 & 5.9\\
            & \mvdustthreer & $\times$ & 18.1 & 5.4 & 3.3 & 18.8 & 6.1 & 2.7 & 17.7 & 5.0 & \textbf{6}3.5\\
            & \mvdustthreerp & $\times$ & \cellcolor[rgb]{0.999, 0.8, 0.8}18.6 & \cellcolor[rgb]{0.999, 0.8, 0.8}5.5 & \cellcolor[rgb]{0.999, 0.8, 0.8}3.0 & \cellcolor[rgb]{0.999, 0.8, 0.8}19.0 & \cellcolor[rgb]{0.999, 0.8, 0.8}6.1 & \cellcolor[rgb]{0.999, 0.8, 0.8}2.6 & \cellcolor[rgb]{0.999, 0.8, 0.8}18.3 & \cellcolor[rgb]{0.999, 0.8, 0.8}5.2 & \cellcolor[rgb]{0.999, 0.8, 0.8}3.1\\\cline{2-12}
            &  {\color{lightgray}$\text{\mvdustthreer}_\text{oracle}$} & $\times$  & \lgr{18.9} & \lgr{5.6} & \lgr{2.8} & \lgr{19.8} & \lgr{6.4} & \lgr{2.3} & \lgr{18.4} & \lgr{5.3} & \lgr{3.0}\\
            &  {\color{lightgray}$\text{\mvdustthreerp}_\text{oracle}$} & $\times$  & \lgr{19.4} & \lgr{5.7} & \lgr{2.6} & \lgr{19.9} & \lgr{6.4} & \lgr{2.3} & \lgr{18.9} & \lgr{5.4} & \lgr{2.8}\\\midrule 
            \multirow{5}{*}{\rotatebox[origin=c]{90}{24 views}} & \dustthreer & \checkmark & 14.3 & 4.2 & 5.6 & 15.2 & 5.0 & 4.2 & 13.8 & 3.6 & 6.1\\
            & \mvdustthreer & $\times$ & 17.8 & 5.3 & 3.6 & 18.4 & 6.0 & 2.9 & 17.3 & 4.9 & 3.8\\
            & \mvdustthreerp & $\times$ & \cellcolor[rgb]{0.999, 0.8, 0.8}18.4 & \cellcolor[rgb]{0.999, 0.8, 0.8}5.4 & \cellcolor[rgb]{0.999, 0.8, 0.8}3.2 & \cellcolor[rgb]{0.999, 0.8, 0.8}18.6 & \cellcolor[rgb]{0.999, 0.8, 0.8}6.0 & \cellcolor[rgb]{0.999, 0.8, 0.8}2.8 & \cellcolor[rgb]{0.999, 0.8, 0.8}17.9 & \cellcolor[rgb]{0.999, 0.8, 0.8}5.1 & \cellcolor[rgb]{0.999, 0.8, 0.8}3.5\\\cline{2-12}
            &  {\color{lightgray}$\text{\mvdustthreer}_\text{oracle}$} & $\times$  & \lgr{18.5} & \lgr{5.5} & \lgr{3.1} & \lgr{19.3} & \lgr{6.2} & \lgr{2.5} & \lgr{18.0} & \lgr{5.1} & \lgr{3.3}\\
            &  {\color{lightgray}$\text{\mvdustthreerp}_\text{oracle}$} & $\times$  & \lgr{19.0} & \lgr{5.6} & \lgr{2.9} & \lgr{19.4} & \lgr{6.2} & \lgr{2.5} & \lgr{18.5} & \lgr{5.3} & \lgr{3.1}\\\bottomrule
        \end{tabular}
    }
    \caption{\textbf{Additional Novel View Synthesis results.}}
    \label{tab:supp_NVS}
\end{table}

\noindent\textbf{\dustthreer baseline}. The \dustthreer-based baseline uses manually defined per-pixel Gaussian parameters. We use the pointmap predicted by \dustthreer as the Gaussian center, use pixel RGB color $I^v$ as the color, a constant $0.001$ for the scale factor $S^{v,m}$, an identity transform, 1.0 for opacity, and spherical harmonics with degree zero. The scale constant is carefully tuned to  balance between  holes and  blurry results. We also remove $3\%$ of the lower confidence points to make the rendering quality as compelling as possible. This is a representative baseline given many pose-free reconstruction methods~\citep{instantsplat, Reconx} are based on \dustthreer with global optimization. To render new views in a relative-scale world space, we follow \cref{sec:nvs_supp} in scaling the scene.

\noindent\textbf{Additional Results.} In~\cref{tab:supp_NVS}, we report additional NVS results varying the number of input views, including 8, 16, and 20 views, which are not reported in Table 4 of the main paper. As more views are used and the scene gets larger, the performance of all methods decreases. However, \mvdustthreerp performs the best in all settings on all 3 evaluation datasets, while \mvdustthreer consistently outperforms the \dustthreer based baseline by a large margin but moderately underperforms \mvdustthreerp.

\subsection{Study: \mvdustthreerp 2-Stage Training}
\label{sec:supp_train_recipe}

\begin{table*}[t]
    \centering

    \resizebox{1.0\columnwidth}{!}{
    \setlength{\tabcolsep}{2pt}
        \begin{tabular}{c|c|c|ccc|ccc|ccc}
            \multirow{2}{*}{Dataset} & \# of  & \multirow{2}{*}{Training recipe}  & \multicolumn{3}{c|}{MVS Reconstruction} & \multicolumn{3}{c|}{MVPE} & \multicolumn{3}{c}{NVS} \\\cline{4-12}
             & views &  & ND $\downarrow$ & DAc $\uparrow$ & CD $\downarrow$ & RRE $\downarrow$ & RTE $\downarrow$ & mAE $\downarrow$ & PSNR $\uparrow$ & SSIM $\uparrow$ & LPIPS $\downarrow$ \\
              \specialrule{.2em}{.1em}{.1em}

\multirow{12}{*}{HM3D} & \multirow{2}{*}{\rotatebox[origin=c]{0}{4}} & 1-stage, 8 views & 1.0 & 95.2 & 1.5(0.9) & 1.2 & 1.1 & 4.9 & 20.2 & 6.1 & 1.9\\
& & 2-stage, mixed views & \cellcolor[rgb]{0.999, 0.8, 0.8}0.9 & \cellcolor[rgb]{0.999, 0.8, 0.8}95.5 & \cellcolor[rgb]{0.999, 0.8, 0.8}1.5(0.7) & \cellcolor[rgb]{0.999, 0.8, 0.8}0.7 & \cellcolor[rgb]{0.999, 0.8, 0.8}0.6 & \cellcolor[rgb]{0.999, 0.8, 0.8}1.9 & \cellcolor[rgb]{0.999, 0.8, 0.8}20.7 & \cellcolor[rgb]{0.999, 0.8, 0.8}6.3 & \cellcolor[rgb]{0.999, 0.8, 0.8}1.8\\\cline{2-12}
& \multirow{2}{*}{\rotatebox[origin=c]{0}{8}} & 1-stage, 8 views & 1.1 & 93.5 & 2.9(0.8) & 0.6 & 0.8 & 2.9 & 19.8 & 6.0 & 2.1\\
& & 2-stage, mixed views & \cellcolor[rgb]{0.999, 0.8, 0.8}1.1 & \cellcolor[rgb]{0.999, 0.8, 0.8}94.0 & \cellcolor[rgb]{0.999, 0.8, 0.8}2.0(1.3) & \cellcolor[rgb]{0.999, 0.8, 0.8}0.5 & \cellcolor[rgb]{0.999, 0.8, 0.8}0.7 & \cellcolor[rgb]{0.999, 0.8, 0.8}2.6 & \cellcolor[rgb]{0.999, 0.8, 0.8}19.9 & \cellcolor[rgb]{0.999, 0.8, 0.8}6.0 & \cellcolor[rgb]{0.999, 0.8, 0.8}2.1\\\cline{2-12}
& \multirow{2}{*}{\rotatebox[origin=c]{0}{12}} & 1-stage, 8 views & 1.2 & 91.5 & 1.8(0.7) & 0.6 & 1.2 & 5.2 & 19.4 & 5.8 & 2.4\\
& & 2-stage, mixed views & \cellcolor[rgb]{0.999, 0.8, 0.8}1.1 & \cellcolor[rgb]{0.999, 0.8, 0.8}93.2 & \cellcolor[rgb]{0.999, 0.8, 0.8}1.5(1.0) & \cellcolor[rgb]{0.999, 0.8, 0.8}0.4 & \cellcolor[rgb]{0.999, 0.8, 0.8}0.7 & \cellcolor[rgb]{0.999, 0.8, 0.8}3.6 & \cellcolor[rgb]{0.999, 0.8, 0.8}19.5 & \cellcolor[rgb]{0.999, 0.8, 0.8}5.9 & \cellcolor[rgb]{0.999, 0.8, 0.8}2.2\\\cline{2-12}
& \multirow{2}{*}{\rotatebox[origin=c]{0}{16}} & 1-stage, 8 views & 1.5 & 85.5 & 2.3(1.5) & 1.0 & 2.5 & 8.5 & 19.0 & 5.6 & 2.7\\
& & 2-stage, mixed views & \cellcolor[rgb]{0.999, 0.8, 0.8}1.3 & \cellcolor[rgb]{0.999, 0.8, 0.8}89.6 & \cellcolor[rgb]{0.999, 0.8, 0.8}2.3(1.3) & \cellcolor[rgb]{0.999, 0.8, 0.8}0.6 & \cellcolor[rgb]{0.999, 0.8, 0.8}1.1 & \cellcolor[rgb]{0.999, 0.8, 0.8}5.0 & \cellcolor[rgb]{0.999, 0.8, 0.8}19.4 & \cellcolor[rgb]{0.999, 0.8, 0.8}5.8 & \cellcolor[rgb]{0.999, 0.8, 0.8}2.4\\\cline{2-12}
& \multirow{2}{*}{\rotatebox[origin=c]{0}{20}} & 1-stage, 8 views & 1.8 & 76.9 & 3.1(1.2) & 1.6 & 4.4 & 12.2 & 18.6 & 5.5 & 3.0\\
& & 2-stage, mixed views & \cellcolor[rgb]{0.999, 0.8, 0.8}1.4 & \cellcolor[rgb]{0.999, 0.8, 0.8}87.0 & \cellcolor[rgb]{0.999, 0.8, 0.8}2.1(1.3) & \cellcolor[rgb]{0.999, 0.8, 0.8}0.8 & \cellcolor[rgb]{0.999, 0.8, 0.8}1.9 & \cellcolor[rgb]{0.999, 0.8, 0.8}6.7 & \cellcolor[rgb]{0.999, 0.8, 0.8}19.2 & \cellcolor[rgb]{0.999, 0.8, 0.8}5.8 & \cellcolor[rgb]{0.999, 0.8, 0.8}2.5\\\cline{2-12}
& \multirow{2}{*}{\rotatebox[origin=c]{0}{24}} & 1-stage, 8 views & 2.1 & 64.5 & 3.9(2.0) & 3.0 & 6.5 & 15.8 & 18.4 & 5.4 & 3.2\\
& & 2-stage, mixed views & \cellcolor[rgb]{0.999, 0.8, 0.8}1.6 & \cellcolor[rgb]{0.999, 0.8, 0.8}83.5 & \cellcolor[rgb]{0.999, 0.8, 0.8}2.4(1.2) & \cellcolor[rgb]{0.999, 0.8, 0.8}1.2 & \cellcolor[rgb]{0.999, 0.8, 0.8}2.7 & \cellcolor[rgb]{0.999, 0.8, 0.8}8.5 & \cellcolor[rgb]{0.999, 0.8, 0.8}19.1 & \cellcolor[rgb]{0.999, 0.8, 0.8}5.7 & \cellcolor[rgb]{0.999, 0.8, 0.8}2.6\\\specialrule{.1em}{.0em}{.0em}

\multirow{12}{*}{ScanNet} & \multirow{2}{*}{\rotatebox[origin=c]{0}{4}} & 1-stage, 8 views & \cellcolor[rgb]{0.999, 0.8, 0.8}0.8 & \cellcolor[rgb]{0.999, 0.8, 0.8}94.9 & \cellcolor[rgb]{0.999, 0.8, 0.8}1.5(0.3) & \cellcolor[rgb]{0.999, 0.8, 0.8}1.4 & 16.1 & 26.2 & 22.2 & 7.1 & 1.5\\
& & 2-stage, mixed views & 0.9 & 94.5 & 1.8(0.3) & 1.5 & \cellcolor[rgb]{0.999, 0.8, 0.8}10.6 & \cellcolor[rgb]{0.999, 0.8, 0.8}20.7 & \cellcolor[rgb]{0.999, 0.8, 0.8}23.1 & \cellcolor[rgb]{0.999, 0.8, 0.8}7.3 & \cellcolor[rgb]{0.999, 0.8, 0.8}1.4\\\cline{2-12} 
& \multirow{2}{*}{\rotatebox[origin=c]{0}{8}} & 1-stage, 8 views & \cellcolor[rgb]{0.999, 0.8, 0.8}1.0 & \cellcolor[rgb]{0.999, 0.8, 0.8}92.1 & \cellcolor[rgb]{0.999, 0.8, 0.8}1.5(0.4) & \cellcolor[rgb]{0.999, 0.8, 0.8}2.0 & \cellcolor[rgb]{0.999, 0.8, 0.8}10.1 & \cellcolor[rgb]{0.999, 0.8, 0.8}21.0 & \cellcolor[rgb]{0.999, 0.8, 0.8}21.3 & \cellcolor[rgb]{0.999, 0.8, 0.8}6.9 & \cellcolor[rgb]{0.999, 0.8, 0.8}1.8\\
& & 2-stage, mixed views & 1.1 & 90.2 & 1.8(0.5) & 2.2 & 10.8 & 21.6 & 21.3 & 6.9 & 1.8\\\cline{2-12} 
& \multirow{2}{*}{\rotatebox[origin=c]{0}{12}} & 1-stage, 8 views & 1.2 & 88.4 & 1.8(0.7) & 2.5 & 11.6 & 22.9 & \cellcolor[rgb]{0.999, 0.8, 0.8}20.4 & 6.6 & 2.1\\
& & 2-stage, mixed views & \cellcolor[rgb]{0.999, 0.8, 0.8}1.2 & \cellcolor[rgb]{0.999, 0.8, 0.8}87.6 & \cellcolor[rgb]{0.999, 0.8, 0.8}1.7(0.8) & \cellcolor[rgb]{0.999, 0.8, 0.8}2.4 & \cellcolor[rgb]{0.999, 0.8, 0.8}11.4 & \cellcolor[rgb]{0.999, 0.8, 0.8}22.7 & 20.3 & \cellcolor[rgb]{0.999, 0.8, 0.8}6.6 & \cellcolor[rgb]{0.999, 0.8, 0.8}2.0\\\cline{2-12} 
& \multirow{2}{*}{\rotatebox[origin=c]{0}{16}} & 1-stage, 8 views & \cellcolor[rgb]{0.999, 0.8, 0.8}1.4 & \cellcolor[rgb]{0.999, 0.8, 0.8}86.6 & \cellcolor[rgb]{0.999, 0.8, 0.8}1.8(0.8) & 3.1 & 13.4 & 25.3 & 19.6 & 6.3 & 2.3\\
& & 2-stage, mixed views & 1.4 & 85.7 & 2.0(0.6) & \cellcolor[rgb]{0.999, 0.8, 0.8}3.0 & \cellcolor[rgb]{0.999, 0.8, 0.8}12.5 & \cellcolor[rgb]{0.999, 0.8, 0.8}24.3 & \cellcolor[rgb]{0.999, 0.8, 0.8}19.6 & \cellcolor[rgb]{0.999, 0.8, 0.8}6.4 & \cellcolor[rgb]{0.999, 0.8, 0.8}2.3\\\cline{2-12} 
& \multirow{2}{*}{\rotatebox[origin=c]{0}{20}} & 1-stage, 8 views & \cellcolor[rgb]{0.999, 0.8, 0.8}1.5 & \cellcolor[rgb]{0.999, 0.8, 0.8}84.0 & \cellcolor[rgb]{0.999, 0.8, 0.8}1.8(0.7) & 3.8 & 15.2 & 27.6 & 19.0 & 6.1 & 2.6\\
& & 2-stage, mixed views & 1.5 & 82.9 & 2.0(0.7) & \cellcolor[rgb]{0.999, 0.8, 0.8}3.8 & \cellcolor[rgb]{0.999, 0.8, 0.8}13.9 & \cellcolor[rgb]{0.999, 0.8, 0.8}26.0 & \cellcolor[rgb]{0.999, 0.8, 0.8}19.1 & \cellcolor[rgb]{0.999, 0.8, 0.8}6.2 & \cellcolor[rgb]{0.999, 0.8, 0.8}2.5\\\cline{2-12} 
& \multirow{2}{*}{\rotatebox[origin=c]{0}{24}} & 1-stage, 8 views & 1.6 & 81.2 & \cellcolor[rgb]{0.999, 0.8, 0.8}1.7(0.7) & 4.6 & 16.7 & 29.4 & 18.6 & 6.0 & 2.8\\
& & 2-stage, mixed views & \cellcolor[rgb]{0.999, 0.8, 0.8}1.5 & \cellcolor[rgb]{0.999, 0.8, 0.8}81.5 & 1.8(0.6) & \cellcolor[rgb]{0.999, 0.8, 0.8}4.5 & \cellcolor[rgb]{0.999, 0.8, 0.8}14.5 & \cellcolor[rgb]{0.999, 0.8, 0.8}27.0 & \cellcolor[rgb]{0.999, 0.8, 0.8}18.7 & \cellcolor[rgb]{0.999, 0.8, 0.8}6.1 & \cellcolor[rgb]{0.999, 0.8, 0.8}2.6\\\specialrule{.1em}{.0em}{.0em}

\multirow{12}{*}{MP3D} & \multirow{2}{*}{\rotatebox[origin=c]{0}{4}} & 1-stage, 8 views & 2.2 & 68.0 & 19.9(3.4) & 0.8 & 0.8 & 4.6 & 19.9 & 5.9 & 2.0\\
& & 2-stage, mixed views & \cellcolor[rgb]{0.999, 0.8, 0.8}2.1 & \cellcolor[rgb]{0.999, 0.8, 0.8}70.4 & \cellcolor[rgb]{0.999, 0.8, 0.8}19.9(3.1) & \cellcolor[rgb]{0.999, 0.8, 0.8}0.8 & \cellcolor[rgb]{0.999, 0.8, 0.8}0.6 & \cellcolor[rgb]{0.999, 0.8, 0.8}2.2 & \cellcolor[rgb]{0.999, 0.8, 0.8}20.4 & \cellcolor[rgb]{0.999, 0.8, 0.8}6.0 & \cellcolor[rgb]{0.999, 0.8, 0.8}1.9\\\cline{2-12} 
& \multirow{2}{*}{\rotatebox[origin=c]{0}{8}} & 1-stage, 8 views & 2.3 & 62.0 & 15.1(4.2) & 0.4 & 0.5 & 2.6 & 19.5 & 5.7 & 2.2\\
& & 2-stage, mixed views & \cellcolor[rgb]{0.999, 0.8, 0.8}2.3 & \cellcolor[rgb]{0.999, 0.8, 0.8}63.4 & \cellcolor[rgb]{0.999, 0.8, 0.8}14.9(3.4) & \cellcolor[rgb]{0.999, 0.8, 0.8}0.4 & \cellcolor[rgb]{0.999, 0.8, 0.8}0.5 & \cellcolor[rgb]{0.999, 0.8, 0.8}2.6 & \cellcolor[rgb]{0.999, 0.8, 0.8}19.5 & \cellcolor[rgb]{0.999, 0.8, 0.8}5.8 & \cellcolor[rgb]{0.999, 0.8, 0.8}2.2\\\cline{2-12} 
& \multirow{2}{*}{\rotatebox[origin=c]{0}{12}} & 1-stage, 8 views & 2.6 & 55.0 & 15.1(3.8) & 0.5 & 1.0 & 4.9 & 19.0 & 5.5 & 2.6\\
& & 2-stage, mixed views & \cellcolor[rgb]{0.999, 0.8, 0.8}2.5 & \cellcolor[rgb]{0.999, 0.8, 0.8}57.2 & \cellcolor[rgb]{0.999, 0.8, 0.8}14.0(2.7) & \cellcolor[rgb]{0.999, 0.8, 0.8}0.5 & \cellcolor[rgb]{0.999, 0.8, 0.8}0.7 & \cellcolor[rgb]{0.999, 0.8, 0.8}3.7 & \cellcolor[rgb]{0.999, 0.8, 0.8}19.1 & \cellcolor[rgb]{0.999, 0.8, 0.8}5.6 & \cellcolor[rgb]{0.999, 0.8, 0.8}2.4\\\cline{2-12} 
& \multirow{2}{*}{\rotatebox[origin=c]{0}{16}} & 1-stage, 8 views & 3.2 & 44.1 & 17.3(4.4) & 1.4 & 2.4 & 8.1 & 18.6 & 5.3 & 2.9\\
& & 2-stage, mixed views & \cellcolor[rgb]{0.999, 0.8, 0.8}2.8 & \cellcolor[rgb]{0.999, 0.8, 0.8}52.0 & \cellcolor[rgb]{0.999, 0.8, 0.8}14.0(3.6) & \cellcolor[rgb]{0.999, 0.8, 0.8}0.8 & \cellcolor[rgb]{0.999, 0.8, 0.8}1.4 & \cellcolor[rgb]{0.999, 0.8, 0.8}5.2 & \cellcolor[rgb]{0.999, 0.8, 0.8}18.9 & \cellcolor[rgb]{0.999, 0.8, 0.8}5.5 & \cellcolor[rgb]{0.999, 0.8, 0.8}2.5\\\cline{2-12} 
& \multirow{2}{*}{\rotatebox[origin=c]{0}{20}} & 1-stage, 8 views & 3.6 & 34.2 & 18.1(6.7) & 1.9 & 3.8 & 11.1 & 18.3 & 5.2 & 3.1\\
& & 2-stage, mixed views & \cellcolor[rgb]{0.999, 0.8, 0.8}3.1 & \cellcolor[rgb]{0.999, 0.8, 0.8}44.8 & \cellcolor[rgb]{0.999, 0.8, 0.8}14.1(3.2) & \cellcolor[rgb]{0.999, 0.8, 0.8}1.0 & \cellcolor[rgb]{0.999, 0.8, 0.8}2.0 & \cellcolor[rgb]{0.999, 0.8, 0.8}6.7 & \cellcolor[rgb]{0.999, 0.8, 0.8}18.7 & \cellcolor[rgb]{0.999, 0.8, 0.8}5.5 & \cellcolor[rgb]{0.999, 0.8, 0.8}2.7\\\cline{2-12} 
& \multirow{2}{*}{\rotatebox[origin=c]{0}{24}} & 1-stage, 8 views & 4.3 & 26.7 & 22.0(5.9) & 3.3 & 6.0 & 14.6 & 17.9 & 5.1 & 3.5\\
& & 2-stage, mixed views & \cellcolor[rgb]{0.999, 0.8, 0.8}3.4 & \cellcolor[rgb]{0.999, 0.8, 0.8}38.1 & \cellcolor[rgb]{0.999, 0.8, 0.8}16.1(3.0) & \cellcolor[rgb]{0.999, 0.8, 0.8}1.5 & \cellcolor[rgb]{0.999, 0.8, 0.8}2.8 & \cellcolor[rgb]{0.999, 0.8, 0.8}8.3 & \cellcolor[rgb]{0.999, 0.8, 0.8}18.5 & \cellcolor[rgb]{0.999, 0.8, 0.8}5.4 & \cellcolor[rgb]{0.999, 0.8, 0.8}2.9\\

             \bottomrule
        \end{tabular}
    }
    \vspace{-0.2cm}
    \caption{\textbf{Comparisons between 1-stage training and 2-stage training}. Models trained with 2 stages perform substantially better on large scenes from HM3D and MP3D, while they perform similarly on small scenes from \scannet.}
    \label{tab:supp_2stage}
\end{table*}

Below we present more ablative studies and a discussion  comparing 1-stage training with fixed 8-view inputs to 2-stage training using a mixed set of inputs with differing numbers of views. As in Section 4.7 of the main paper, in 2-stage training, we finetune the model trained in the 1st stage with fixed 8-view input for another 70 epochs. For this, we randomly sample inputs with a varying number of views $N$ sampled between 4 and 12 views. In~\cref{tab:supp_2stage}, we find the models with 2-stage training consistently outperform the models trained with 1-stage training on datasets of large scenes, including HM3D and MP3D. Models with 2-stage training perform comparably on small scenes in \scannet. The improvements on MP3D are more significant when the number of input views is large (e.g., 24 views). We conclude, the 2nd stage training with mixed input views forces the model to be more robust to scene size and the number of input views. This improves the model's generalization performance on large scenes.

\clearpage
\clearpage


\end{document}